\documentclass[11pt]{scaleai-paper}

\usepackage{amsmath}
\usepackage{amsfonts}
\usepackage{amssymb}
\usepackage{amsthm}
\usepackage{booktabs}
\usepackage{tabularx}
\usepackage{tabulary}
\usepackage{multirow}
\usepackage{subcaption}
\usepackage{float}
\usepackage{graphicx}
\usepackage{import}
\usepackage{fancyvrb}
\usepackage{fvextra}
\usepackage[square,numbers,sort&compress]{natbib}
\usepackage{xspace}
\usepackage{url}
\usepackage[colorlinks=true,linkcolor=scaleLink,citecolor=scaleLink,urlcolor=scaleLink]{hyperref}
\usepackage[capitalise,nameinlink]{cleveref}
\usepackage{fontawesome5}

\newcolumntype{Y}{>{\RaggedRight\arraybackslash}X}
\Crefformat{equation}{#2Equation~#1#3} 
\crefformat{equation}{#2eq.~#1#3}      

\newcommand{\method}{CoverSynth}

\papertype{Scale AI Research}
\contact{\textcolor{black}{\faEnvelope} \quad \texttt{anahita.sharma@scale.com}}

\title{CoverSynth: Coverage-Guided Generation of Retrieval Evaluation Sets}

\author[1]{Anahita Sharma}
\author[1]{Apaar Shanker}
\author[1]{Rishav Chakravarti}
\author[1]{Sneha Srinivasan}
\author[1]{Vijay Kalmath}
\author[1]{Dylan Wilbur}
\author[1]{Yuan Li}
\author[1]{Sara Bolouki}
\author[1]{Yuan Xue}
\affil[1]{Scale AI}

\begin{document}

\maketitle

\begin{abstract}
Enterprise agents need retrieval evaluation sets that expose domain-specific failures and guide developers towards efficient deployment tradeoffs. Neither expert-human-authored sets nor current synthetic generation methods provide a methodology for achieving the necessary coverage, reliability, efficiency, and discriminability required for such decision-making. Simultaneously, the field does not have an assessment framework for determining when an evaluation set has satisfied these properties.  We introduce an evaluation set assessment framework for these criteria and \method\ -- a generator specifically designed to satisfy these criteria without incurring the high cost of human expert labeling. Starting from an unannotated corpus, \method\ targets coverage across retrieval demands and system capabilities. Deterministic checks locate evidence quotes in their sources, while model-based checks assess semantic validity to guarantee reliable ground truth. Across five public enterprise-style corpora, our method provides informative variation in difficulty, revealing unseen failure modes such as an inability of agents to handle underspecified or ambiguous requests. On the RepLiQA corpus, 91\% of human-generated queries can be answered by all members on a panel of 12 retrieval systems, compared to only 12\% of \method{}-generated queries. Finally, in a repeated TechQA/RepLiQA study, 12 of 12 within-harness model contrasts have positive nominal 95\% intervals on generated sets, compared with 4 of 12 on human sets. \method{} makes coverage and evidence quality explicit construction objectives, producing corpus-grounded evaluations that reveal retrieval headroom and differences between agent configurations.

\end{abstract}

\section{Introduction}
Enterprise agents operate over large, heterogeneous private corpora with organization-specific terminology, document relationships, and operational rules \citep{enterpriseragbench2026}. Beyond filling gaps in pre-trained knowledge \citep{kandpal2023longtail}, agents must interpret information in its local context \citep{agarwal2026tribal}, and when retrieved evidence drives decisions, missing dependencies or unsupported assumptions propagate through later workflow steps. Evaluation must therefore test both whether an agent can assemble supporting evidence and whether it recognizes requests the corpus cannot resolve.

An industry survey finds that only 5\% of organizations have a custom generative AI tool in production, attributing most failures to brittle workflows and a lack of contextual learning \citep{challapally2025genaidivide}. Retrieval alone does not guarantee grounding: retrieval-based legal research tools hallucinate on 17--33\% of queries \citep{magesh2024hallucinationfree}, and a search-augmented model leaves about 30\% of its medical statements unsupported by its cited sources \citep{wu2025sourcecheckup}. Regulators also increasingly require pre-deployment testing of high-risk AI systems \citep[e.g.,][]{euaiact2024}.

Enterprises often need these evaluations without existing questions, gold annotations or query logs. Authoring by subject-matter experts (SMEs) is costly: ViDoRe~V3 required 12,000 hours of annotation \citep{vidore2026}; confidentiality also limits crowd annotation of regulated corpora \citep{cook2025fintechrag}. Recent benchmarks combine model drafting with human or LLM-judge verification \citep{enterpriseragbench2026,wixqa2025,vidore2026}, or automate generation as in MiRAGE \citep{mirage2026}. Verification of individual items, however, does not establish coverage of an intended workload. Our five-corpus profiles show that native benchmark and MiRAGE reference sets leave 31--59\% of concept clusters untested (\Cref{tab:quality}). The two MiRAGE reference sets contain only answerable questions, leaving abstention and clarification behavior untested.

Such omissions can conceal failures on particular retrieval demands; incomplete gold and questions on which systems saturate can also obscure useful performance differences. We translate these risks into an assessment framework for evaluation sets (\Cref{sec:assessment}). Coverage and diversity metrics quantify concentration across question types and evidence structures, representation of user roles, and untested corpus regions \citep{Wang2026,coverage2026}. Source-localization checks and oracle evidence probes address annotation weaknesses that can distort retrieval scores. Measurements of headroom, repeat-based system separation and resource use assess whether the set can inform choices among systems and budgets. These complementary diagnostics apply across construction methods and are interpreted relative to the declared workload and evidence representation.

Guided by these criteria, we present \method, a framework that synthesizes retrieval evaluation sets from an unannotated corpus (\Cref{fig:system}). It makes coverage across corpus regions, personas, intents, evidence structures and failure modes an explicit construction objective. Deterministic checks locate retained evidence quotes exactly in their sources, while model-based checks assess semantic validity. The pipeline infers lenses and personas and requires no expert-authored questions or evidence annotations; expert effort can instead focus on corpus preparation, reviewing the intended workload and auditing outputs (\Cref{sec:protocol}).

We use this assessment framework to evaluate \method\ against native and synthetic references on five enterprise-style corpora (\Cref{sec:results}). The reported profiles show higher joint attribute diversity (0.77--0.91 vs.\ 0.35--0.70) and no uncovered documents on three corpora. On RepLiQA, all 12 tested systems cite every gold document in every available repeat on 91\% of human questions versus 12\% of generated questions. In the repeated TechQA/RepLiQA study, 12 of 12 within-harness model contrasts have positive nominal 95\% intervals on generated sets, compared with 4 of 12 on human sets. Answerability verdicts further expose difficulty recognizing underspecified and ambiguous requests.

Our contributions are:
\begin{itemize}
    \item \textbf{An assessment framework for retrieval evaluation sets}, mapping practical requirements for coverage, evidence validity, and system comparison to complementary, automatically computed diagnostics (\Cref{sec:assessment}).
    \item \textbf{A method for constructing retrieval evaluation sets directly from unannotated corpora}, with verifiable evidence annotations and explicit coverage objectives, evaluated against native and synthetic references across five corpora (\Cref{sec:method} and \Cref{sec:results}).
    \item \textbf{A taxonomy and controlled generation procedure for unanswerable queries}, covering missing evidence, false premises, under-specification, and ambiguity. By applying controlled changes to answerable questions, it creates cases that test whether agents recognize when a request cannot be answered as posed (\Cref{sec:negatives}).
\end{itemize}

\section{Related Work}
\label{sec:related}

\paragraph{Generation of a retrieval evaluation set.}
Enterprise retrieval benchmarks such as EnterpriseRAG-Bench \citep{enterpriseragbench2026}, WixQA \citep{wixqa2025}, and ViDoRe~V3 \citep{vidore2026} rely on human authoring or curation, and report that different question types need different construction procedures and that negative coverage requires deliberate corpus exploration. \method\ reproduces these properties synthetically.

InPars \citep{inpars2022} and Promptagator \citep{promptagator2022} introduced synthetic query generation via domain-agnostic and few-shot prompting. Later work added control and verification: attribute guidance with programmatic checks (DataGen \cite{datagen2024}), multi-agent verifiers (MiRAGE \cite{mirage2026}), and chunk-level knowledge graphs for single- and multi-hop synthesis (RARE \cite{rare2025}; RAGAS \cite{ragas2024}). Other frameworks target visual page retrieval \citep{vidore2026}, preference arenas (DocArena \cite{docarena2026}), or multi-step reasoning trajectories (DRAGON \cite{dragon2025}; S$^3$-R1 \cite{Goel2026}). None guarantees byte-exact evidence by construction or isolates the sequential dependency depth that iterative multi-hop retrieval requires.

\paragraph{Coverage of the corpus and of question types.}
\citet{coverage2026} show that aggregate metrics hide localized failures and propose semantic stratification to cover the whole corpus. CCQGen \citep{ccqgen2025} conditions generation on uncovered concepts, and OBLIQ-Bench \citep{oblique2026} clusters documents by a latent attribute before writing queries. \method\ also clusters the corpus, but targets joint coverage of question types, answerability classes, and evidence structure rather than topics alone.

\paragraph{Unanswerability and abstention.}
RefusalBench \citep{refusalbench2025} categorizes informational uncertainty as ambiguity, contradiction, missing information, or false premise; HopRefusalBench \citep{hoprefusalbench2026} shows that where unanswerability occurs in a reasoning chain (root, intermediate, or terminal) affects refusal behavior; and GENERATE-PAIRWISE \citep{itsallrelative2023} jointly synthesizes relevant and irrelevant queries as hard negatives. \method\ turns these findings into a generation procedure: it applies one controlled edit to a verified answerable anchor and keeps the item only if the corpus supports the intended failure, fixing the locus and cause of unanswerability by construction.

\paragraph{Assessing synthetic data quality.}
SynQuE \citep{synque2025} scores synthetic data against unlabeled samples using embedding distribution and diversity metrics plus LLM-based structural rubrics (LENS). Our framework instead measures intrinsic properties of the evaluation set and needs no reference data.

\Cref{tab:positioning} (\Cref{app:related-works}) compares \method\ with representative methods.

\section{Method}
\label{sec:method}

\begin{figure}[t]
\begin{center}
\includegraphics[width=0.85\linewidth]{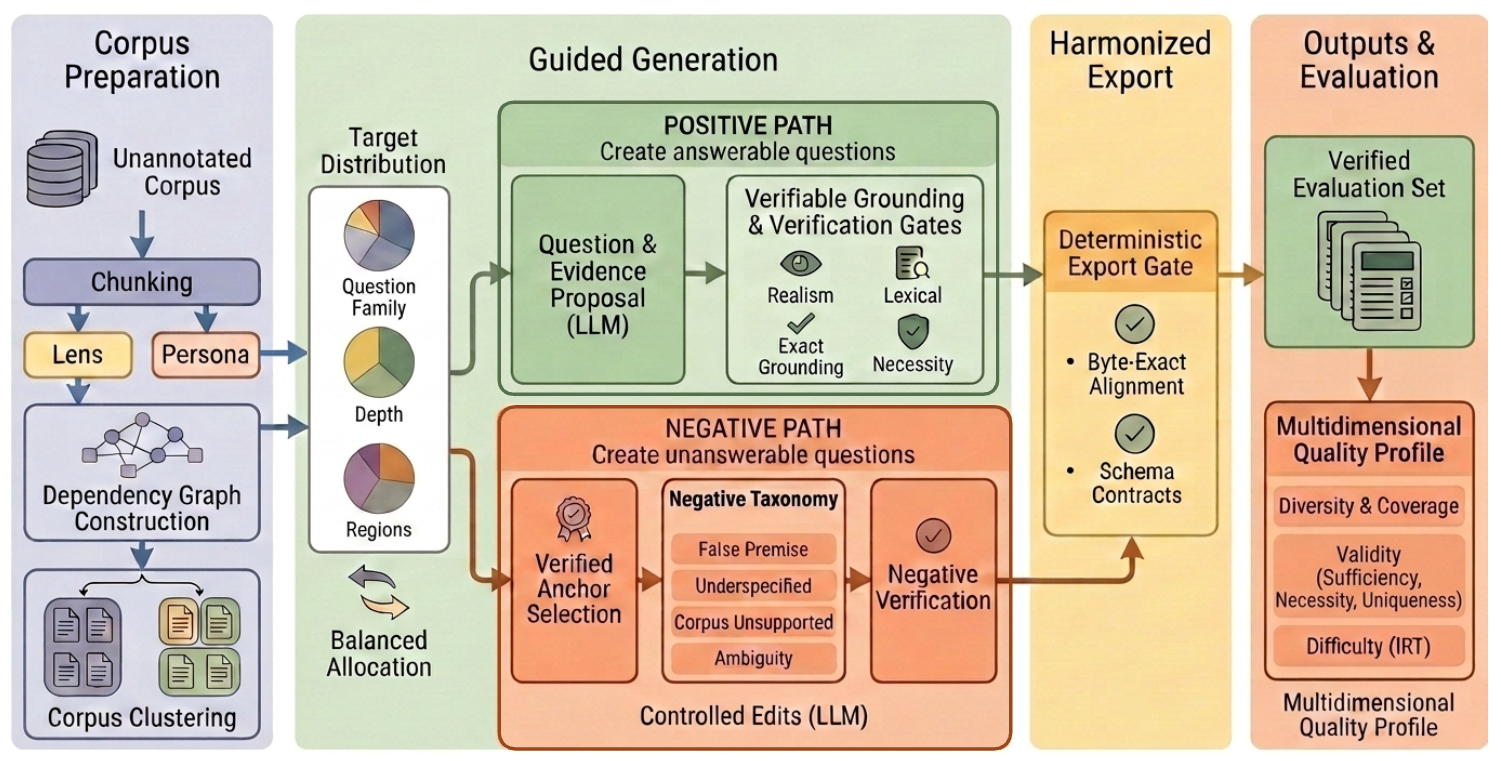}
\end{center}
\caption{Overview of \method. An inferred lens guides chunk annotation and concept-graph construction. Graph communities, personas, and coverage controls guide positive and negative query generation. Model checks assess semantic validity; the export gate enforces exact span localization and record contracts.}
\label{fig:system}
\end{figure}

\method\ constructs an evaluation set from an unannotated corpus $\mathcal{C}$. Models propose lenses, personas, questions, and quotes while keeping track of question attributes that support non-redundant coverage as well as precise downstream error analysis. Deterministic components localize quotes, validate schemas, and analyze the graph. Positive and negative synthesis paths meet at an independent export gate (\Cref{fig:system}).

\subsection{Corpus representation and dependency graph}
\label{sec:clustering}
Layout-aware parsing splits documents into chunks, from which models infer one \emph{lens} and a small set of \emph{user personas}. Inspired by OBLIQ-Bench's human-defined lens \citep{oblique2026}, the lens specifies how to interpret passages, such as the engineering requirement each establishes. Personas describe expected users, their domain familiarity, question style, and example questions.

The lens guides annotation of concepts, references, and acronyms from each chunk. Canonicalized concept labels form the vertices $V=\{\ell_1,\dots,\ell_M\}$. Semantic edges connect each concept to its $k=15$ nearest neighbors with cosine similarity at least $\tau=0.55$, using $256$-dimensional \texttt{text-embedding-3-large} embeddings. Citation edges connect concepts whose passages cross-reference one another, with fixed weight $0.55$. Resulting in a graph $G=(V,E_{\text{sem}}\cup E_{\text{cite}})$

By partitioning the concept graph into localized communities, Leiden clustering maps each chunk to its concepts' communities, grouping related passages across the corpus. Within each community, a planner classifies passage subsets as \emph{co-dependent} (2--4 jointly required passages), \emph{competing} (opposing or conditionally bounded claims), or \emph{isolated} (single-passage facts) to identify feasible question types. 

\subsection{Query topology: depth versus breadth}
\label{sec:topology}
For evidence set $S=\{s_1,\dots,s_b\}$, breadth $b=|S|$ counts evidence units and depth $d$ counts sequential dependencies:
\begin{itemize}
    \item \textbf{Single lookup ($b=1, d=0$):} One evidence unit supplies the answer.
    \item \textbf{Parallel aggregation ($b>1, d=0$):} Independent units jointly supply the answer.
    \item \textbf{Bridge chain ($b>1, d=b-1$):} Resolving $s_t$ yields the key needed to locate $s_{t+1}$.
\end{itemize}
Only bridges contribute to hop depth; parallel aggregation increases breadth.

\subsection{Verifiable ground-truth spans}
\label{sec:invariant}
To construct answerable evaluation items, the model drafts a question
from a feasible passage subset (\Cref{sec:clustering}) and proposes
verbatim supporting quotes. Each quote $s_k\in S$ is drawn from a source
chunk $c_k$. A deterministic locator checks each quote against its source
and records the matching character offsets.

A quote is admitted only if it occurs in its source:
\begin{equation}
g_{\text{ground}}(s_k, c_k) \;=\; \mathbf{1}\!\left[\,\exists\,[i, j]\ \text{s.t.}\ c_k[i : j] = s_k\,\right].
\label{eq:ground}
\end{equation}
The matching offsets $[i_k,j_k]$ are stored in the item record. Every cited span must pass this check during generation and independently at export, including after negative generation (\Cref{sec:negatives}) as well.  

\subsection{Positive generation and adversarial pruning}
\label{sec:positives}
A draft uses $\mathcal{T}=(\text{structure},\,\text{persona},\,\text{intent},\,\text{steer})$: the query topology, a persona, one of eight intents (\Cref{tab:intents}), and a bias toward under-produced difficulty and phrasing, respectively. Ordered checks impose two hard gates: persona realism ($g_{\text{real}}$; \Cref{app:realism_prompt}) and exact quote localization ($g_{\text{ground}}$; \Cref{eq:ground}). Two soft checks prompt repairs for question--passage bag-of-words overlap above $0.5$ and distinctive wording or a question that lists
its own answer options (the semantic ``own-words'' critic; \Cref{app: own_words_prompt}). A final adversarial verifier ($g_{\text{verify}}$) checks answerability, gold-span sufficiency, that the question neither
reveals the answer nor points to its location, independent support per evidence unit, and sequential dependency for bridges (\Cref{app:single_adversarial_prompt}--\ref{app:bridge_adversarial_prompt}). The hard admission criteria are
\begin{equation}
A(q,\{s_k\}) \;=\; g_{\text{real}} \;\wedge\; \left(\bigwedge_{k=1}^{|S|} g_{\text{ground}}(s_k, c_k)\right) \;\wedge\; g_{\text{verify}}.
\label{eq:admission}
\end{equation}
A draft still failing a hard gate after at most two repairs is rejected. An accepted positive becomes a \emph{verified anchor}.

\subsection{Negative generation and difficulty}
\label{sec:negatives}
Each negative applies one generation-control edit (\Cref{tab:taxonomy}) to a verified anchor. Two separate axes target refusal difficulty: \emph{topology} places the defect at the root, intermediate, or terminal hop, leaving preceding hops answerable \citep{hoprefusalbench2026}; \emph{intensity} (low, medium, high) sets the variation's severity and intended detectability \citep{refusalbench2025}. Two independently framed adjudicators assess validity. A \emph{direct verifier} checks the intended failure, such as corpus silence, a false premise, or reference ambiguity; an \emph{adversarial solver} tries to resolve it against the corpus. Admission requires the verifier to confirm the failure and the solver to find no resolution.

\subsection{Coverage control}
\label{sec:coverage}
Positive candidates are enumerated over the graph under fixed per-source caps. A round-robin schedule prioritizes under-sampled documents; an equal-share hop quota controls depth. Soft steering discourages over-produced phrasing and complexity tiers. Stable hashing of candidate tuples assigns valid persona--intent pairs, with realized coverage assessed post-hoc (\Cref{sec:assessment}).

For negatives, each attribute axis---failure family, generation control, topology, intensity, and complexity---is apportioned separately. Given normalized target weights $w$ and budget $n$, largest-remainder (Hamilton) apportionment assigns
\begin{equation}
n_j = \lfloor n\,w_j \rfloor + r_j, \qquad \sum_j n_j = n,
\label{eq:apportion}
\end{equation}
where $r_j\in\{0,1\}$ assigns remaining slots to the largest fractional parts. Greedy emission minimizes the maximum relative deficit, $\max_j (n\,w_j-\hat n_j)/(n\,w_j)$, among values feasible for the active anchor; $\hat n_j$ counts emitted or reserved items. Marginal apportionment avoids enumerating the full Cartesian grid, but anchor feasibility can still constrain realized coverage.

\section{Assessment Framework}
\label{sec:assessment}

Retrieval evaluation should help developers diagnose failures, compare candidate systems, and make informed deployment decisions for an intended workload. An evaluation set that omits relevant information needs or corpus regions can leave important failures untested. Unreliable reference evidence can confound system errors with annotation artifacts, while questions on which candidate systems uniformly succeed or fail offer little signal for choosing between them. These risks motivate three complementary requirements: coverage of retrieval demands within the intended workload, validity of gold evidence, and informative variation in difficulty. Our assessment framework operationalizes these requirements through automated diagnostics, interpreted relative to a declared corpus and workload.

\paragraph{Intended users and scope.}
Personas specify the user roles and information needs an enterprise evaluation targets. Coverage is assessed within this declared scope. For a declared persona set $\mathcal{P}$ with question counts $q_h$ and shares $\pi_h = q_h / \sum_{h'\in\mathcal{P}} q_{h'}$, persona coverage is the normalized entropy
\begin{equation}
H_{\mathrm{norm}} \;=\; -\frac{\sum_{h\in\mathcal{P}} \pi_h \log \pi_h}{\log |\mathcal{P}|} \;\in\; [0,1],
\label{eq:persona}
\end{equation}
which equals 1 when every persona receives the same number of questions and 0 when all questions target a single persona. Questions that match no declared persona are excluded from $\pi_h$.

\paragraph{Coverage and diversity.}
Retrieval performance can vary with both question demands and corpus regions: success on direct lookup need not transfer to cross-file aggregation or sequential reference resolution, and omitted topics leave local failures untested.
We assess eight question intents (\Cref{app:taxonomy-tables}) jointly with multi-file and multi-hop requirements. For the declared feasible cells $\Omega$ of this capability grid, let $p$ be the empirical item distribution. Attribute diversity \citep{Wang2026} is the normalized joint entropy
\begin{equation}
\mathrm{AD} \;=\; -\frac{\sum_{\omega\in\Omega} p(\omega)\,\log p(\omega)}{\log|\Omega|} \;\in\; [0,1],
\label{eq:ad}
\end{equation}
which captures interaction imbalances missed by marginal entropies.
Following \citet{coverage2026}, we measure corpus coverage with the zero-query cluster proportion ($\mathrm{ZQC}$) and sufficient-corpus-coverage ratio ($\mathrm{SCC}$). For corpus document set $\mathcal{C}$, assessment clusters $\mathcal{K}$, and queries $Q_i$ whose gold evidence lies in a document with a chunk that belongs to cluster $i$,
\begin{equation}
\mathrm{ZQC} \;=\; \frac{\big|\{i \in \mathcal{K} : |Q_i| = 0\}\big|}{|\mathcal{K}|} \;\in\; [0,1], \qquad
\mathrm{SCC} \;=\; \frac{|\mathcal{D}_{\text{covered}}|}{|\mathcal{C}|},
\label{eq:corpuscov}
\end{equation}
where $\mathcal{D}_{\text{covered}}$ contains documents in at least one cluster with $|Q_i|\ge n_{\min}=5$. A document is assigned to every cluster that any of its chunks joins, so one document typically belongs to several clusters. ZQC identifies clusters receiving no questions, serving as a coverage floor; SCC measures document coverage through repeatedly queried clusters. Taken together, $AD$, $ZQC$, and $SCC$ measure a benchmark's ability to detect performance variances across dimensions which matter for retrieval.

\paragraph{Gold-evidence validity.}
Incomplete or redundant gold can confound system errors with annotation artifacts. Exact localization (\Cref{eq:ground}) checks that evidence exists; oracle-reader probes assess its semantic usefulness. \emph{Sufficiency} is the fraction of probed questions answered correctly from gold alone. Conditional on gold sufficiency, \emph{uniqueness} is the fraction answered incorrectly from selected distractors, and \emph{necessity} is the fraction of gold units whose removal makes the reader incorrect. These probes depend on the reader and evidence granularity; results appear in \Cref{tab:spanvalidity}, with formal definitions in \Cref{app:oracle-probes}.

\paragraph{Difficulty.}
An evaluation set should exercise routine and demanding cases, revealing failures and where additional capability may help. We examine document-retrieval saturation and the stability of system differences across repeats as evidence of diagnostic value within the tested panel (\Cref{sec:results};  \Cref{app:historical-irt} and \ref{app:focused-protocol}).

\section{Experimental Setup}
\label{sec:setup}

\subsection{Corpus selection and preparation}
\label{sec:corpora}
We evaluate \method\ on five public collections with varied retrieval demands. TechQA contains short, sectioned IBM support technotes; RepLiQA contains fictional prose. EnterpriseRAG-Bench (ERAG-Bench) spans nine source systems---Confluence, Jira, Slack, GitHub, Google Drive, Gmail, Linear, HubSpot, and Fireflies---with a Confluence subset used for generation. eCFR contains nested, cross-referenced regulations. Harvey Firm Knowledge (Harvey FK) contains legal documents organized by client, where near-duplicate content across clients makes it easy to ground an answer in the wrong client's records.

We convert corpora to Markdown and chunk each document along its native structure (headings,
capitalized section labels, or numbered clauses), falling back to paragraph blocks where none exist. Subsets approximate deployment scope: a Confluence space, an organization's policies, whole legal matters, or a set of regulators. Gold-seeded TechQA and ERAG subsets contain a $0.40$ document fraction of distractors. We inspect chunks-per-document distributions and resample degenerate subsets, including those with large single-chunk tails. Each corpus is capped at 5{,}000 chunks to limit generation cost. Sampling choices are fixed before generation. Detailed corpus sub-sampling methodology can be found in \Cref{app:corpora-sampling}.

\subsection{Evaluation protocol and baselines}
\label{sec:protocol}
\paragraph{Generation details.} 
GPT-5.4 annotates concepts, entities, and references; GPT-5.6 Sol drafts questions and adjudicates semantics. Lens and personas are inferred, with target hop depths of 0, 1, or 2.

\paragraph{Retrieval systems and studies.} Our nine-system base panel combines BM25, dense retrieval (\texttt{all-MiniLM-L6-v2}), and their reciprocal-rank fusion with ReAct and Claude Code, each using Haiku-4.5, Sonnet-5, and Opus-5. Agents iteratively search, inspect documents, and select evidence; returned spans are scored for precision, recall, and $F_2$ (\Cref{app:retrieval-config}). Single-pass Claude Code measurements across the five corpora appear in \Cref{app:original-retrieval}.

\paragraph{Repeated two-corpus study.} On TechQA and RepLiQA, whose human reference sets provide passage-level gold, we run the nine-system panel on both the generated and the human sets, repeating each agent three times; indexed retrieval is deterministic. This study examines headroom and model separation (\Cref{sec:focused-results}). Settings and scoring populations are in \Cref{app:focused-protocol}, and additional single-run agent configurations in \Cref{app:extended-configs}.

\paragraph{Baselines.}
We use native ERAG, TechQA and RepLiQA references, and MiRAGE \citep{mirage2026} for Harvey and eCFR. MiRAGE uses chunk-level gold, scored separately from spans (\Cref{app:original-retrieval}).

\section{Results and Discussion}
\label{sec:results}
\label{sec:discussion}

We first assess the composition, evidence validity, and retrieval behavior of the generated sets across all five corpora. A focused repeated study on TechQA and RepLiQA then examines measurement headroom and system separation against human sets with passage-level gold.

\subsection{Coverage and evidence validity}
\label{sec:coverage-results}

\begin{table}[t]
\caption{Benchmark assessment metrics as defined in \Cref{sec:assessment} on the five corpora using both \method{} as well as reference benchmarks. Note that persona entropy values are calculated using CoverSynth's persona set, excluding questions without a matching persona. Computation of reference datasets' SCC and ZQC are detailed in \Cref{app:clustering}. Arrows indicate the preferred direction for the first four assessment metrics. For difficulty skew, values closer to zero are preferred (with negative values representing a skew towards hard and positive values representing a skew towards easy). Statistically significant differences for AD, persona coverage, and difficulty skew are marked with $^{*}$.}
\label{tab:quality}
\centering
\small
\setlength{\tabcolsep}{4pt}
\begin{tabular}{llrccccc}
\toprule
Corpus & Source & Queries & AD $\uparrow$ & Persona $H_{\mathrm{norm}}$ $\uparrow$ & ZQC $\downarrow$ & SCC $\uparrow$ & Difficulty skew $\odot$ \\
\midrule
\multirow{2}{*}{ERAG-Bench} & \method & 150 & \textbf{0.88}$^*$ & \textbf{0.98}$^*$ & \textbf{0.00} & \textbf{0.96} & $-0.69$ \\
 & Reference & 53 & 0.70 & 0.79 & 0.54 & 0.86 & $-0.90$ \\
\midrule
\multirow{2}{*}{Harvey FK} & \method & 150 & \textbf{0.89}$^*$ & \textbf{0.99}$^*$ & \textbf{0.00} & 0.76 & $-0.67^{*}$ \\
 & Reference & 150 & 0.59 & 0.87 & 0.59 & \textbf{0.99} & $-1.60$ \\
\midrule
\multirow{2}{*}{TechQA} & \method & 128 & \textbf{0.91}$^*$ & \textbf{0.99}$^*$ & \textbf{0.18} & \textbf{0.87} & $+0.27$ \\
 & Reference & 374 & 0.41 & 0.54 & 0.31 & 0.54 & $-0.09$ \\
\midrule
\multirow{2}{*}{eCFR} & \method & 148 & \textbf{0.84}$^*$ & \textbf{0.99}$^*$ & \textbf{0.00} & \textbf{1.00} & $-0.29$ \\
 & Reference & 150 & 0.66 & 0.76 & 0.44 & 0.28 & $-0.43$ \\
\midrule
\multirow{2}{*}{RepLiQA} & \method & 147 & \textbf{0.77}$^*$ & \textbf{0.99}$^*$ & \textbf{0.02} & \textbf{0.98} & $-0.36^{*}$ \\
 & Reference & 685 & 0.35 & 0.79 & 0.57 & 0.67 & $+2.31$ \\
\bottomrule
\end{tabular}
\end{table}

\paragraph{Broader coverage, with corpus-dependent tradeoffs.}
\method\ achieves higher joint attribute diversity than references across all corpora (\Cref{tab:quality}). ZQC is lower in every comparison and is reported as zero on ERAG-Bench, Harvey and eCFR. SCC improves on ERAG-Bench, TechQA, eCFR and RepLiQA, but is lower on Harvey (0.76 versus 0.99). Reaching an assessment cluster at least once therefore differs from meeting the repeated-coverage threshold in clusters associated with a document. Retroactively labeling reference queries with CoverSynth's persona set reveals that CoverSynth yields near-uniform persona coverage, while reference benchmarks achieve significantly lower spread. These profiles support broader measured coverage while exposing how repeated coverage is distributed under the specified inputs.

\begin{table}[t]
\caption{Gold-span validity, as defined in \Cref{sec:assessment}, from oracle-reader probes. The oracle reader (Claude-Opus-5) is a different model from the generator (GPT-5.6 Sol). Uniqueness contrasts each gold span against the three highest lexical-overlap non-gold passages. $^{\dagger}$The human TechQA and RepLiQA sets carry a single gold span per item, so leave-one-out necessity is trivially $1.00$ and not comparable to the multi-span \method\ rate.}
\label{tab:spanvalidity}
\begin{center}
{\small
\setlength{\tabcolsep}{4pt}
\resizebox{\linewidth}{!}{%
\begin{tabular}{l cc cc cc cc cc}
\toprule
& \multicolumn{2}{c}{\textbf{ERAG-Bench}} & \multicolumn{2}{c}{\textbf{Harvey FK}} & \multicolumn{2}{c}{\textbf{TechQA}} & \multicolumn{2}{c}{\textbf{eCFR}} & \multicolumn{2}{c}{\textbf{RepLiQA}} \\
\cmidrule(lr){2-3} \cmidrule(lr){4-5} \cmidrule(lr){6-7} \cmidrule(lr){8-9} \cmidrule(lr){10-11}
\textbf{Rate} & \textbf{CoverSynth} & \textbf{Human} & \textbf{CoverSynth} & \textbf{MiRAGE} & \textbf{CoverSynth} & \textbf{Human} & \textbf{CoverSynth} & \textbf{MiRAGE} & \textbf{CoverSynth} & \textbf{Human} \\
\midrule
Sufficiency               & \textbf{0.70} & 0.64 & 0.80 & \textbf{0.95} & \textbf{0.72} & 0.61 & \textbf{0.97}  & 0.95 & \textbf{0.91} & 0.90 \\
Uniqueness                & 0.99 & \textbf{1.00} & \textbf{0.68} & 0.65 & 0.96 & 0.96 & \textbf{0.99} & 0.94 & 0.90 & \textbf{0.97} \\
Necessity (leave-one-out) & 0.58 & \textbf{0.64} & \textbf{0.61} & 0.24 & 0.67 & ---$^{\dagger}$ & \textbf{0.49} & 0.41 & 0.49 & ---$^{\dagger}$ \\
\bottomrule
\end{tabular}%
 }
}
\end{center}
\end{table}

\paragraph{Exact localization and semantic validity.}
Source-offset checks inherently ensure exact localization of evidence span offsets. In addition, \Cref{tab:spanvalidity}'s oracle-reader probes show that  \method\ achieves sufficiency of 0.70--0.97, uniqueness of 0.68--0.99, and leave-one-out necessity of 0.49--0.67 scores. These indicate higher validity scores on five of six quality dimensions relative to MiRAGE, i.e., the other synthetically generated baseline. However, this only remains true for half the dimensions when comparing to human reference baselines. This suggests that human reference baselines may yield more precise evidence spans, offering room for improvement by \method{}.

\subsection{Retrieval behavior and negative-query diagnostics}
\label{sec:retrieval-results}

\paragraph{Retrieval agent discriminability}
In \Cref{fig:focused-system-scores} we see that both human and \method{} discriminate reliably between agentic vs. indexed retrievers in span $F_2$ (with the indexed retrievers reliably performing worse than agentic retrievers). However, only \method{} can consistently discern the expected Haiku-4.5 $<$ Sonnet-5 $<$ Opus-5 base model hierarchy. Further, in \Cref{fig:five-corpus-measurements} (\Cref{app:original-retrieval}), \method{} is able to discern each harness type by the expected context token consumption rate (unlike the reference benchmarks). Complete per-model measurements and protocol details appear in \Cref{app:original-retrieval}.

\begin{figure}[t]
\centering
\includegraphics[width=\linewidth]{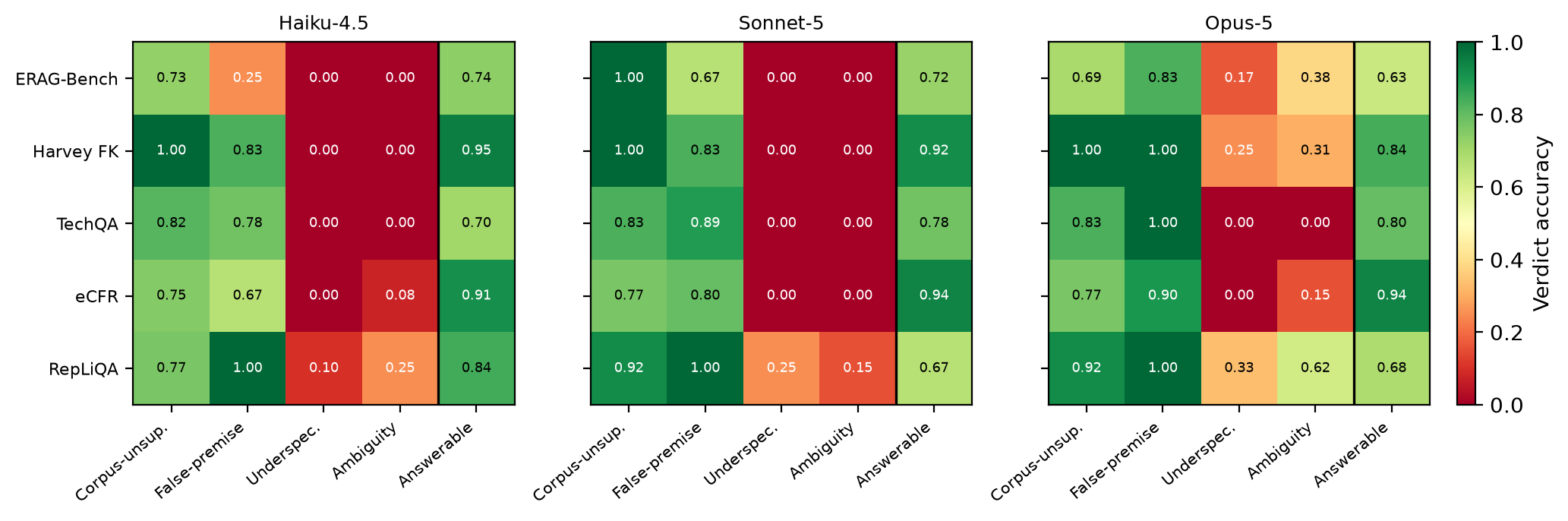}
\caption{Single-pass answerability-verdict accuracy by corpus and model under the Claude Code harness, for all five \method{}  outputs (the same runs as \Cref{tab:downstream}). The first four columns are the negative families; the final column is answerable-item verdict accuracy. Failed runs are excluded. }
\label{fig:negatives}
\end{figure}

\paragraph{Negative families expose different failure modes.}
In \Cref{fig:negatives}, we see that \method{} not only discerns the expected model backbone hierarchy, but reveals the specific failure modes (underspecified and ambiguous questions) that the stronger reasoning model (Opus-5) is most effective at improving upon. The taxonomy therefore reveals distinctions obscured by aggregate answerability accuracy. See \Cref{app:negative-results} for the detailed breakout.

\subsection{Performance headroom and retrieval system separation}
\label{sec:focused-results}

We next examine whether the generated sets support informative system comparisons on TechQA and RepLiQA, using the repeated nine-system panel described in \Cref{sec:protocol}.

\begin{figure}[t]
\centering
\includegraphics[width=0.8\linewidth]{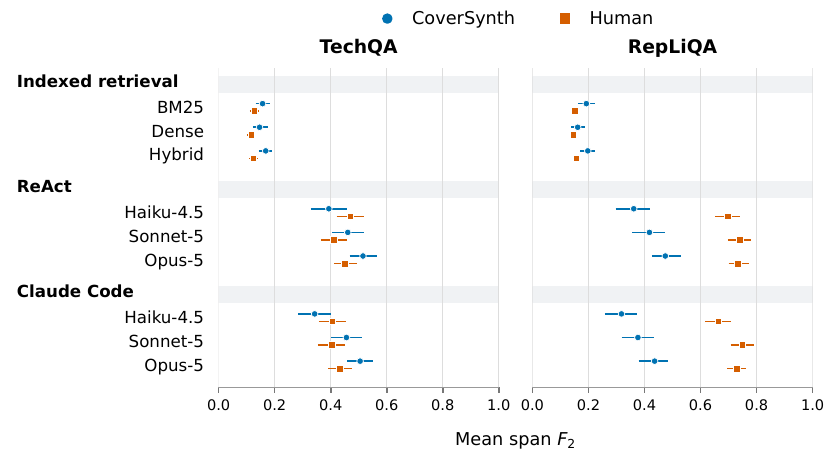}
\caption{Retrieval performance on \method\ and human evaluation sets. Points show mean span $F_2$ on answerable questions; whiskers show 95\% bootstrap intervals. Protocol details are in \Cref{app:focused-protocol}.}
\label{fig:focused-system-scores}
\end{figure}

\paragraph{Performance varies by corpus.}
\Cref{fig:focused-system-scores} shows all configurations separately. Agents score lower on generated RepLiQA questions, while TechQA differences vary by model. Within the base ReAct and Claude Code harnesses, 12 of 12 ordered model contrasts have positive nominal 95\% intervals on generated sets, compared with 4 of 12 on human sets. Contrasts are oriented according to the expected Haiku--Sonnet--Opus capability order. Full scores and model contrasts are in \Cref{app:focused-protocol}.

\paragraph{Headroom and reliability.}
In \Cref{tab:quality}'s difficulty skewness metric (also visualized in \Cref{fig:difficulty-distribution} of \Cref{app:historical-irt}), we see that \method{} produces a less skewed distribution of difficulty levels on four out of the five corpora as compared to their respective reference baselines. In particular, it avoids the extreme skew towards hard questions that MiRAGE creates on Harvey FK (a skew of only -0.67 by \method{} vs. MiRAGE's -1.60) and the extreme skew towards easy questions that RepLiQA ships with (a skew of of only -0.36 by \method{} vs. the reference benchmark's +2.31). This difficulty distribution directly translates into more reliable discernment of different retrieval systems. For example, 91\% of the answerable subset of benchmark queries from the native RepLiQA corpus are correctly addressed by every member of the nine of panel of retrieval systems. Making these retrieval systems effectively indistinguishable on the native benchmark. This is only true for 12\% of \method{}'s generated queries, allowing for more informative deployment choices.

\subsection{Discussion}
\label{sec:results-implications}

Measured against the assessment framework outlined in \Cref{sec:assessment}, we show that \method{} consistently provides more coverage than what native and MiRAGE references provide on all five corpora. This includes more extensive retrieval demands (including underspecified and ambiguous demands), better distribution over the domain's natural personas as measured by entropy, as well as less skewed difficulty distributions. Persona coverage is computed against \method{}'s corpus-derived persona sets, and evaluating those sets against a more comprehensive topic structure is left to future work. The focused study adds further evidence of head room and model separation not only on span~$F_2$, but also along other performance dimensions such as token consumption. The rich taxonomy supported by \method{} also allows drill downs into specific failure modes that explain the separation between systems not only at an aggregate score level, but also at fine grained retrieval demands (such as underspecified intents). Finally, this greater coverage and discernment is accompanied by high gold label validity scores relative to the other synthetic baseline benchmark, making the resulting automated corpus driven benchmark an effective tool for high quality deployment decision making in enterprise scenarios.

\section{Conclusion}
\label{sec:conclusion}

We introduced an assessment framework for retrieval evaluation sets and \method, a generator that
builds such sets from an unannotated corpus with coverage and exact evidence localization as
construction objectives. On five enterprise-style corpora, the resulting sets cover the declared
workload more broadly than native and MiRAGE references and separate retrieval systems where human
sets saturate.

\section*{AI Use Statement}
We used a generative AI assistant to draft and edit the prose of this manuscript, to format its equations and tables, and to organize the bibliography. The method, the mathematical formulation, the experimental design, and all reported results are the authors' own work; the implementation was completed with the assistance of AI coding tools. The authors reviewed and verified all text, claims, and citations, and take responsibility for the final content.

\section*{Ethics Statement}
Our method generates evaluation data from public corpora only, all under permissive or public-domain licenses. Grounding every item in verifiable evidence reduces the risk of unsupported claims in downstream measurement. We use no proprietary or personal data in this work.

\bibliographystyle{abbrvnat}
\bibliography{references}

\appendix
\section{Appendix}

\subsection{Comparison to existing methodologies for evaluation set generation}
\label[appendix]{app:related-works}

\begin{table}[H]
\caption{Positioning of \method\ against representative synthetic generation methods (references in \Cref{sec:related}) on five dimensions: reliance on human annotation or guidance (SME-free), validation of byte-exact text spans (grounding), inclusion of multi-hop and negative questions (multi-hop and negative, respectively), and coverage-aware generation (coverage). \checkmark: supported; $\sim$: partial or heuristic; --: not addressed. Each prior method covers a subset of these properties; \method\ is distinguished by combining byte-exact grounding with a formal structural approach to generating multi-hop positives and hard negatives under enforced coverage.}
\label{tab:positioning}
\begin{center}
{\small
\begin{tabular}{l ccccc}
\toprule
\textbf{Method} & \textbf{SME-free} & \textbf{Grounding} & \textbf{Multi-hop} & \textbf{Negatives} & \textbf{Coverage} \\
\midrule
InPars              & \checkmark & --         & --         & --         & --         \\
Promptagator        & $\sim$     & --         & --         & --         & --         \\
DataGen             & \checkmark & $\sim$     & --         & --         & $\sim$     \\
RARE                & \checkmark & $\sim$     & \checkmark & --         & --         \\
CCQGen              & \checkmark & --         & --         & --         & \checkmark \\
OBLIQ-Bench         & $\sim$     & $\sim$     & --         & --         & \checkmark \\
EnterpriseRAG-Bench & --         & \checkmark & $\sim$     & \checkmark & $\sim$     \\
MiRAGE              & \checkmark & $\sim$     & --         & --         & $\sim$     \\
DocArena            & \checkmark & $\sim$     & $\sim$     & --         & --         \\
RefusalBench        & $\sim$         & \checkmark & --         & \checkmark & --         \\
HopRefusalBench     & $\sim$         & \checkmark & \checkmark & \checkmark & --         \\
\midrule
\textbf{\method\ (ours)} & \checkmark & \checkmark & \checkmark & \checkmark & \checkmark \\
\bottomrule
\end{tabular}
}
\end{center}
\end{table}

\subsection{Corpora Sampling}
\label[appendix]{app:corpora-sampling}

\paragraph{Baseline domain filtering.} Each collection is first restricted to the sub-domain that matches the material enterprise retrieval agents are actually deployed over, rather than to the incidental bulk of the collection. EnterpriseRAG-Bench is filtered on (\texttt{source\_type}) to Confluence: wiki pages hold the documented, canonical knowledge — runbooks, specifications, internal policy — that an enterprise knowledge agent is typically pointed at, whereas the systems that dominate by volume, Slack and Gmail, are transient conversational traffic. RepLiQA is filtered on (\texttt{document\_topic}) to Company Policies, the one topic of seventeen that is workplace material rather than local-news reportage, with the filter applied before sampling so the document budget draws within the topic. Harvey contributes the legal knowledge-work setting and is sampled at the level of whole matters rather than files, because a matter is the unit its gold addresses and a partial matter cannot answer a question about it. eCFR contributes the compliance setting, cut to five titles — banking, securities, food and drug, labor, and public welfare — fixed in advance as the regulators a compliance-facing retrieval agent would plausibly be pointed at. TechQA needs no filter: every document is an IBM product support technote, already the enterprise support scenario.

\paragraph{Gold-first seeding and distractors.} Where a collection ships human questions, the corpus is seeded with the documents its gold cites rather than taken as an arbitrary slice — an arbitrary slice would leave most human questions unanswerable against the prepared corpus, confounding any generated-versus-human comparison with missing evidence. TechQA, EnterpriseRAG-Bench and RepLiQA are each then filled out with a sampled distractor set held at a matched fraction of 0.40 of the corpus, so that retrieval difficulty is comparable across the three. Harvey takes no separate draw, being sampled by whole matter, and eCFR ships no questions at all, so in both the remainder of the corpus serves as the distractor pool for any given question. Selection is seeded throughout.

\paragraph{Conversion and segmentation.} No collection ships in a format the generator reads natively, so each is converted to Markdown and segmented into chunks — the unit over which gold spans and multi-chunk questions are built. Segmentation uses whatever structure the source provides: markdown headings for Confluence pages, capitalised section labels for TechQA technotes, and numbered clauses for eCFR, whose sections are individually chunk-sized and are therefore aggregated upward into parts so that a document is a complete regulation rather than a fragment. Documents with no recoverable structure fall through to a paragraph-block segmenter. This step is verified rather than assumed: a document that yields no segmentation collapses to a single whole-file chunk, which removes within-document questions silently rather than failing, so we check the realized chunk distribution per corpus. All corpora but RepLiQA segment more than 98\% of their documents, at medians of 6 to 22 chunks. RepLiQA averages 1.9, its text being PDF-derived prose with no internal structure, so we treat it as a single-chunk corpus and report within-document breadth only for the others.

\subsection{Question intents and negative taxonomy}
\label[appendix]{app:taxonomy-tables}

\begin{table}[H]
\caption{The eight question intents (\texttt{scenario\_type}) a positive item can realize, each naming the analytical work the answer requires. The generator steers a candidate toward one intent and an adversarial gate confirms the drafted question matches it. \emph{Factual lookup} is the default, used when no other intent fits.}
\label{tab:intents}
\begin{center}
{\small

\begin{tabular}{p{2.9cm} p{9.4cm}}
\toprule
\textbf{Question intent} & \textbf{What the answer requires} \\
\midrule
Factual lookup & Retrieve a stated fact, definition, value, threshold, name, or reference precisely, as written (the default when no other intent fits). \\[3pt]
Compliance & Whether a given situation satisfies a rule, or what is required to satisfy it: eligibility, permissibility, exemption, or which requirement governs the case. \\[3pt]
Procedural & How something is done or constructed; the answer is a method, a sequence of steps, or a process. \\[3pt]
Diagnostic & Explains an existing state: the reason or mechanism behind a rule or behavior, or where a defect or failure lies. \\[3pt]
Impact & What follows from a state or change: its effect, consequence, penalty, or knock-on obligation. \\[3pt]
Temporal & Turns on time or order: what must happen before or after what, a stage, duration, waiting period, frequency, or deadline; or how guidance on a topic changed across dated sources. \\[3pt]
Comparative & How two concepts relate, differ, or compare, on an axis other than time or source agreement. \\[3pt]
Discrepancy resolution & Whether the sources agree: reconciling an inconsistency, deciding which of two positions governs, or confirming they align; requires passages that genuinely diverge. \\
\bottomrule
\end{tabular}
}
\end{center}
\end{table}

\begin{table}[H]
\caption{The negative-question taxonomy: four failure families realized by eight generation controls. Answerability and expected behavior are determined by the family. Only the corpus-unsupported family ships an empty gold set; the others cite the passage that makes the item fail.}
\label{tab:taxonomy}
\begin{center}
{\small
\begin{tabular}{p{2.7cm} p{3.7cm} p{2.3cm} p{3.7cm}}
\toprule
\textbf{Failure family} & \textbf{Generation control(s)} & \textbf{Answerability} & \textbf{Desired agent behavior} \\
\midrule
Corpus-unsupported & Entity, property, condition, time, scope & Unsupported & Abstain with rationale \\[4pt]
False premise & Contradicted assertion & Unsupported & Correct the premise \\[4pt]
Underspecified & Remove disambiguator & Ambiguous & Request clarification \\[4pt]
Ambiguity & Ambiguous reference & Ambiguous & Request clarification \\
\bottomrule
\end{tabular}
}
\end{center}
\end{table}

\subsection{Verification Prompt: Realism gate (\texorpdfstring{$g_{\text{real}}$}{g\_real})}
\label[appendix]{app:realism_prompt}
\begin{Verbatim}[breaklines=true,breakanywhere=true,fontsize=\footnotesize]
You are {persona}.

The question below was written for a file-search assistant that has NOT seen any source text and must
SEARCH a corpus to answer it. Make TWO independent judgments.

1. self_contained - is the question about the SUBJECT itself rather than the PAPERWORK? Naming a
   SPECIFIC, real-world entity the question is ABOUT is FINE and often realistic - a court case, statute
   or section citation, ruling, party, product, or a named standard/spec/identifier a practitioner would
   cite ("In [Case v. Party], what did the court hold about [issue]?" is self_contained=TRUE:
   the case is the subject). Mark self_contained=FALSE only if, in ANY phrasing, it:
   - refers to the SOURCE ARTIFACT or how it is organized - "the document / guide / manual / section /
     table / annex", a GENERIC "the standard" / "the guidance" / "the local practice", "which section
     covers X", "where is X documented", "point me to...";
   - asks what a document SAYS instead of the substantive answer (really "where is this written?");
   - refers deictically to an unseen passage ("the scenario described", "as shown", "used here",
     "listed here", "the provided material");
   - only makes sense if the reader is already looking at the document.
   The line: naming the thing the answer is ABOUT (a named case, statute, spec, product, or procedure) =
   subject = TRUE; pointing at the file/section it lives in, or a *generic/unnamed* "the document /
   the standard" = paperwork = FALSE.

2. realistic - would YOU genuinely ask this in day-to-day work? Mark realistic=FALSE if it is
   over-specified, telegraphs its own answer, pre-lists or enumerates the answer's parts (e.g.
   "(a)...(b)...", "list each...", "...including X, Y and Z", "such as ..."), or is longer / more precise than a
   busy person would type. (Two closely-related clauses like "what does X mean and how is it used?" are
   fine.)

QUESTION: {question}

Respond ONLY JSON: {"self_contained": true|false, "realistic": true|false,
  "reason": "<short - which check failed and why>",
  "suggestion": "<a self-contained, natural rephrasing that asks about the subject>"}
\end{Verbatim}

\subsection{Verification Prompt: Own-words critic (lexical gate)}
\label[appendix]{app: own_words_prompt}
\begin{Verbatim}[breaklines=true,breakanywhere=true,fontsize=\footnotesize]
You are auditing whether a retrieval question is too EASY - answerable without really
searching or reasoning over the material. You are shown the question and the EXACT source passage(s) it
is meant to be answered from.

Mark searchable=true if EITHER of these holds:
A. KEYWORD LIFT - the question reuses the passage's distinctive wording (its rare multi-word technical
   phrases or sentence structure), so a reader could find the passage by keyword search without
   understanding the concept.
B. ANSWER TELEGRAPHED - the question STATES or LISTS the answer itself: it names the specific value(s),
   option(s), or alternatives that ARE the answer, so the agent only has to CONFIRM a choice rather than
   find and reason it out. Examples:
   - "...do I use the household-income test or the individual-income test?" when those two ARE the answer
     -> ask "...which income test applies here?"
   - "...do I just carry over the unadjusted baseline rate?" when that IS the answer
     -> ask "...what rate basis do I use for this case?"
   - "...is it 14 days or 30 days?" when the value is the answer -> ask "...what filing deadline applies?"

Mark searchable=false if answering requires understanding: the question asks about the same underlying
point in DIFFERENT words / framing than the passage AND does not hand over the answer's values/options.

IMPORTANT: do NOT penalize ordinary domain vocabulary a real user would naturally use (a term of art,
the standard name for a thing). Penalize only DISTINCTIVE lifted phrasing or a telegraphed answer. A
question can be perfectly natural and domain-specific while still being open and in the asker's own words.

QUESTION: {question}

SOURCE PASSAGE(S):
{quotes}

Respond ONLY JSON: {"searchable": true|false, "reason": "<short: 'lift' or 'telegraphs answer' + why>",
  "suggestion": "<the same question re-asked openly, in the asker's own words, WITHOUT the answer in it>"}
\end{Verbatim}

\subsection{Adversarial Verifier Prompt: Single and Aggregation Queries}
\label[appendix]{app:single_adversarial_prompt}
\begin{Verbatim}[breaklines=true,breakanywhere=true,fontsize=\footnotesize]
You are an adversarial grader checking a retrieval-eval item.

QUESTION: {question}

EVIDENCE QUOTES (verbatim from the corpus; each is a separate piece of evidence, id in brackets):
{quotes}

Decide, using ONLY these quotes:
1. answerable: can the question be answered correctly from these quotes alone?
2. sufficient: is the evidence complete (nothing essential missing)?
3. per_doc: for EACH evidence id, does that quote INDEPENDENTLY support the part of the answer it
   contributes? (catches a "multi-hop" question that is really answerable from a single piece of
   evidence - including two passages from the SAME document.)
4. necessary: for EACH quote in order, is it actually needed (true) or redundant/spurious (false)?
5. leak: does the QUESTION give away the answer, or point at WHERE in the source it is
   written (a line or page number, "see the table below", "in section 4 of the guide")? Naming a
   real-world entity the question is ABOUT - a case, a statute or section citation, a named
   standard - is NOT a leak; that is how a practitioner asks.{intent_criterion}

Respond with ONLY JSON:
{"answerable": bool, "sufficient": bool, "per_doc": {"<evidence_id>": bool},
  "necessary": [bool, ...], "leak": bool,{intent_field} "notes": "<short>"}
\end{Verbatim}

\subsection{Adversarial Verifier Prompt: Bridge (multi-hop) Queries}
\label[appendix]{app:bridge_adversarial_prompt}
\begin{Verbatim}[breaklines=true,breakanywhere=true,fontsize=\footnotesize]
You are an adversarial grader checking a MULTI-HOP (scavenger-hunt) retrieval item.
The evidence quotes are given IN INTENDED CHAIN ORDER: each earlier quote should name a "bridge" entity
you must resolve before you can even locate the next quote.

QUESTION: {question}

EVIDENCE QUOTES (verbatim, in chain order, id in brackets):
{quotes}

Bridge entities the question must NOT name (naming one lets the agent skip a hop): {bridges}

Decide, using ONLY these quotes:
1. answerable: can the question be answered correctly from these quotes together?
2. sufficient: is the evidence complete (nothing essential missing)?
3. dependency: is at least one OTHER quote genuinely REQUIRED to reach the answer - either because it
   supplies a distinct necessary fact, OR because it DECODES a term/acronym or IDENTIFIES an entity the
   question uses that you must resolve to interpret the question or LOCATE the answer? Answer TRUE if
   removing that quote would leave a searcher unable to correctly identify, interpret, or find the
   answer - EVEN IF the quote that states the answer reads as self-contained once you are looking at it.
   Answer FALSE only if every other quote is genuinely redundant (e.g. the answer quote already uses the
   resolved/expanded form, so nothing needed decoding, and no other quote is needed to locate it).
   Example of a REAL dependency: the question uses an acronym; quote A spells it out; quote B states the
   rule using the full form - B alone never tells you what the acronym means, so A is required.
4. bridge_leak: does the QUESTION hand over the RESOLVED form of a linking entity - its expansion /
   definition, or the specific intermediate identity - so the reader can skip a hop and jump straight to
   the answer? (Merely USING an acronym or an unresolved term is FINE - that IS the hop; giving its full
   form or definition is the leak.)
5. necessary: for EACH quote in order, is it actually needed to traverse the chain (true) or
   redundant/spurious (false)? An intermediate quote whose removal still leaves the chain resolvable is
   NOT necessary.
6. leak: does the QUESTION give away the answer, or point at WHERE in the source it is
   written (a line or page number, "see the table below", "in section 4 of the guide")? Naming a
   real-world entity the question is ABOUT - a case, a statute or section citation, a named
   standard - is NOT a leak; that is how a practitioner asks.{intent_criterion}

Respond with ONLY JSON:
{"answerable": bool, "sufficient": bool, "dependency": bool, "bridge_leak": bool,
  "necessary": [bool, ...], "leak": bool,{intent_field} "notes": "<short>"}
\end{Verbatim}

\subsection{Retrieval System Configurations}
\label[appendix]{app:retrieval-config}

\paragraph{Indexed baselines.} BM25, dense retrieval, and hybrid retrieval share a common passage segmentation: documents are split into passages of at most 250 tokens using the \texttt{cl100k\_base} tokenizer, with up to 50 tokens of overlap. Each baseline returns up to 10 ranked passages as evidence. BM25 uses the BM25S implementation with Lucene scoring ($k_1=0.9$, $b=0.4$), lowercasing, English stopword removal, and Snowball English stemming. Dense retrieval uses \texttt{all-MiniLM-L6-v2} embeddings.

\paragraph{Hybrid retrieval.} Hybrid retrieval retrieves up to 20 candidates from each of the same BM25 and dense retrievers, then combines their rankings using equal-weight reciprocal-rank fusion with a rank constant of 60. A passage at rank $r$ contributes $1/(60+r)$ to its fused score from each candidate list in which it appears, with ranks starting at one. The top 10 passages under the fused ranking are returned without an additional reranking stage.

\paragraph{Agent harnesses and budgets.} The local ReAct and Claude Code harnesses each run with Haiku-4.5, Sonnet-5, and Opus-5, yielding six agent configurations. Agents iteratively search and inspect the corpus and select supporting passages as evidence. Agent runs are capped at 50 turns and a 300\,s wall-clock timeout, with one retry permitted for rate-limit errors.

\subsection{Construction profiles and exact localization}
\label{app:construction-results}

\Cref{tab:clustering} retains the corpus-specific document and community counts. Community counts depend on corpus size, the concept representation, and clustering settings; they are not a controlled measure of domain complexity. \Cref{tab:realized-structure} reports realized properties of the current generated TechQA/RepLiQA files. Multi-document counts use distinct gold source files; multi-hop counts use the recorded dependency-depth labels and do not reestablish the semantic necessity of every bridge.

\begin{table}[H]
\caption{Reported Leiden community counts (communities with at least two concept labels) on the corpus subsets used for the construction profiles.}
\label{tab:clustering}
\centering
\small
\begin{tabular}{lrr}
\toprule
Corpus & Documents & Generation Communities \\
\midrule
ERAG-Bench & 95 & 95 \\
Harvey FK & 137 & 515 \\
TechQA & 500 & 176 \\
eCFR & 120 & 398 \\
RepLiQA & 275 & 65 \\
\bottomrule
\end{tabular}
\end{table}

\begin{table}[H]
\centering
\small
\caption{CoverSynth-generated set composition. Structural counts and AD use the 100 answerable items per corpus; intent presence uses all records. AD is normalized over the declared 32-cell grid. Negative-family counts refer to distinct items, not repeated runs.}
\label{tab:realized-structure}
\begin{tabular}{lrr}
\toprule
Property & TechQA & RepLiQA \\
\midrule
All items & 128 & 147 \\
Answerable items & 100 & 100 \\
Multi-document answerable items & 50 & 72 \\
Labeled multi-hop answerable items & 34 & 11 \\
Occupied capability cells & 28/32 & 19/32 \\
AD & 0.906 & 0.768 \\
Intents present & 7/8 & 7/8 \\
\midrule
Corpus-unsupported negatives & 12 & 13 \\
False-premise negatives & 9 & 9 \\
Underspecified negatives & 5 & 12 \\
Ambiguous negatives & 2 & 13 \\
\bottomrule
\end{tabular}
\end{table}

The offset check covers every gold quote in the four current TechQA/RepLiQA files, including evidence records attached to negatives: 244 generated TechQA quotes, 405 generated RepLiQA quotes, and 280 human-reference quotes, for 929 matches with no missing sources or mismatched substrings. This establishes occurrence at the stored character offsets. It does not test whether an answer is correct, whether a passage is necessary, or whether a query is representative of user traffic.

The human ERAG retrieval subset contains 73 items, including 53 answerable items; 43 of those answerable items cite multiple documents. Consequently, the single-document structure of the human TechQA/RepLiQA subsets cannot be generalized to human retrieval benchmarks as a whole.

\subsection{Coverage profiling}
\label[appendix]{app:clustering}
The MiRAGE ZQC/SCC profiles on Harvey and eCFR use the implementation accompanying \citet{coverage2026}. These measure question coverage, independently of the chunk retrieval scores below. The assessment partition $\mathcal{K}$ in \Cref{sec:assessment} need not be the graph partition used for generation. Interpreting the profiles requires a common document inventory, assessment partition, and query-assignment rule within each corpus comparison.

\subsection{Difficulty analysis.}
\label[appendix]{app:historical-irt}
For the historical backbone panel, we fit a two-parameter logistic item response model. System $a$ succeeds on item $i$ with probability
\begin{equation}
P(Y_{ai}=1 \mid \theta_a,\alpha_i,\beta_i) \;=\; \sigma\!\left(\alpha_i(\theta_a - \beta_i)\right) \;=\; \frac{1}{1+\exp\!\left(-\alpha_i(\theta_a - \beta_i)\right)},
\label{eq:irt}
\end{equation}
where $\theta_a$ is latent system ability, $\beta_i$ item difficulty, and $\alpha_i$ item discrimination. Maximum a posteriori estimation uses a standard-normal ability prior and a zero-mean ridge prior on $(\alpha_i,\beta_i)$. Span~$F_2$ is binarized at the panel's mean-matching threshold before fitting.

\begin{figure}[H]
\centering
\includegraphics[width=\linewidth]{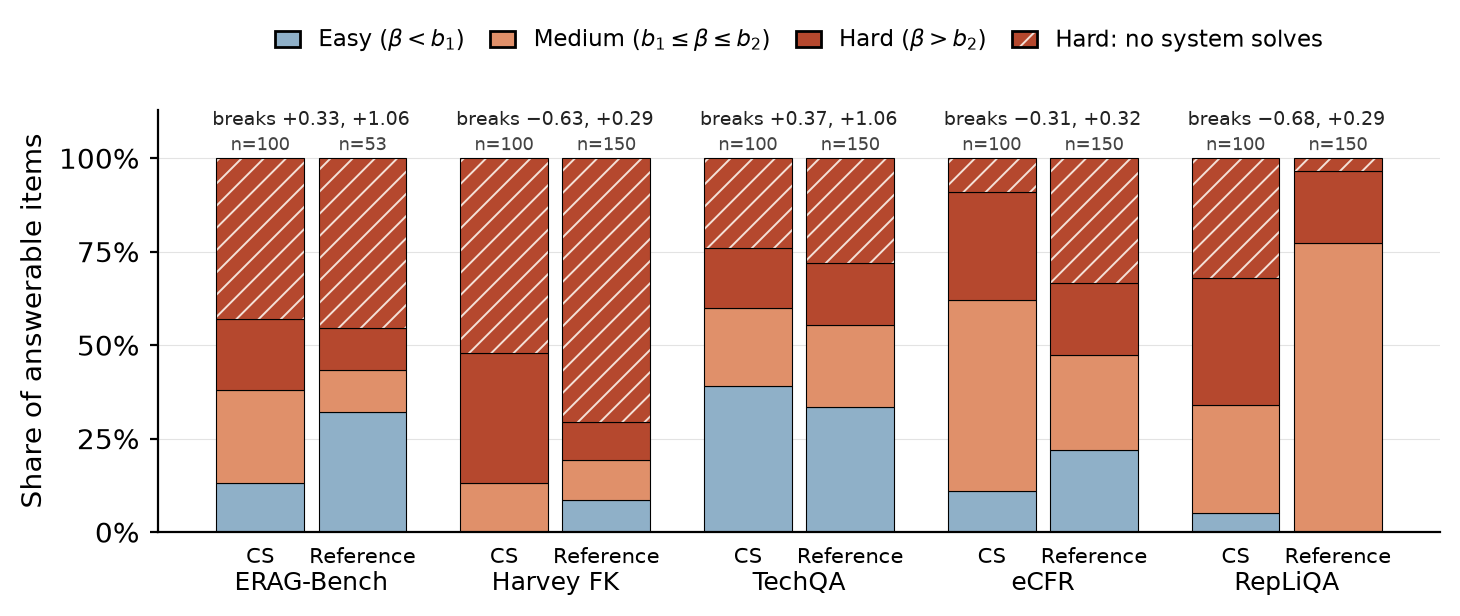}
\caption{Difficulty distribution of answerable questions for \method\ and each reference set. Difficulty is IRT item difficulty $\beta$ from a joint fit per corpus over six systems (three indexed retrievers and three Claude Code agents). Cut-offs for each corpus's difficulty category were determined using Jenks natural breaks optimization. Hatched segments represent the proportion of hard questions that no system was able to solve.}
\label{fig:difficulty-distribution}
\end{figure}

\begin{table}[H]
\centering
\small
\setlength{\tabcolsep}{4pt}
\caption{Shape of the item difficulty distribution on answerable questions. An item's failure rate is the share of the six panel systems whose binarized span-$F_2$ falls below the corpus's pooled mean-matching threshold. Skew is the adjusted Fisher--Pearson sample skewness of item failure rates; positive values indicate mass at the easy end with a tail of harder items. $\Delta$ skew gives \method\ minus \textit{reference} with a 95\% bootstrap interval (2{,}000 resamples, drawing items independently within each set).}
\label{tab:difficulty-skew}
\begin{tabular}{llcc}
\toprule
Corpus & Set & Skew & $\Delta$ skew [95\% CI] \\
\midrule
\multirow{2}{*}{ERAG-Bench} & \method & $-0.69$ & \multirow{2}{*}{$+0.21$ $[-0.54,+0.92]$} \\
 & Reference & $-0.90$ & \\
\midrule
\multirow{2}{*}{Harvey FK} & \method & $-0.67$ & \multirow{2}{*}{$+0.93$ $[+0.37,+1.53]$} \\
 & Reference & $-1.60$ & \\
\midrule
\multirow{2}{*}{TechQA} & \method & $+0.27$ & \multirow{2}{*}{$+0.35$ $[-0.21,+0.90]$} \\
 & Reference & $-0.09$ & \\
\midrule
\multirow{2}{*}{eCFR} & \method & $-0.29$ & \multirow{2}{*}{$+0.14$ $[-0.27,+0.77]$} \\
 & Reference & $-0.43$ & \\
\midrule
\multirow{2}{*}{RepLiQA} & \method & $-0.36$ & \multirow{2}{*}{$-2.67$ $[-3.38,-1.98]$} \\
 & Reference & $+2.31$ & \\
\bottomrule
\end{tabular}
\end{table}

\subsection{Oracle-reader probes and evidence granularity}
\label[appendix]{app:oracle-probes}

\Cref{tab:spanvalidity} reports reader probes, which are distinct from retrieval precision, recall, and $F_1$. The gold evidence units for \method\ are localized spans. Sufficiency depends on the amount of supplied evidence, uniqueness on the chosen distractors, and necessity on the unit removed. A whole-chunk deletion and a localized-quote deletion therefore test different interventions. We do not infer relative evidence quality across unmatched representations. MiRAGE is excluded from these oracle comparisons because its native annotations identify chunks and a matched span-level reader protocol has not been established.

Let the oracle reader's correctness on evidence $S$ be $r(S)\in\{0,1\}$, and let $S_i^*$ be item $i$'s gold evidence units and $D_i$ its distractor passages. Over $N$ probed items, \emph{sufficiency} is the share answered correctly from gold alone: $\mathrm{Sufficiency}=\frac{1}{N}\sum_{i=1}^{N}r(S_i^*)$. Conditional on gold sufficiency, \emph{uniqueness} is the share answered incorrectly from distractors alone: $\mathrm{Uniqueness}=\sum_{i:r(S_i^*)=1}\mathbf{1}[r(D_i)=0]\,/\,\sum_{i=1}^N r(S_i^*)$. \emph{Necessity} is the fraction of gold units in sufficient items whose removal makes the oracle incorrect:
\begin{equation}
\mathrm{Necessity} \;=\; \frac{\sum_{i:\,r(S_i^*)=1}\ \sum_{s \in S_i^*}\mathbf{1}\!\left[\,r(S_i^*\setminus\{s\})=0\,\right]}{\sum_{i:\,r(S_i^*)=1}\,|S_i^*|}
\label{eq:necessity}
\end{equation}

Human TechQA sufficiency (0.61) illustrates that oracle answerability limits also affect human-authored sets. A bridge passage may help an agent locate an answer while being unnecessary to a reader already given the final answer passage; redundant parallel evidence can also lower leave-one-out necessity. These probes are not estimates of human annotation error or proofs of global answer uniqueness.

\subsection{Five-corpus retrieval measurements}
\label[appendix]{app:original-retrieval}

Tables~\ref{tab:downstream}, \ref{tab:baseline} and~\ref{tab:mirage-chunk-retrieval} report retrieval performance on answerable questions from \method, native benchmark references and MiRAGE, respectively, using Claude Code with Opus-5, Sonnet-5 and Haiku-4.5. The generated and reference arms were run separately, with turn budgets of 50 (\method) and 40 (native references and MiRAGE). All three tables exclude failed runs. In the \method\ arm, Haiku excludes 12 failed answerable runs on TechQA, 10 on RepLiQA, 9 on ERAG-Bench, 6 on eCFR and one on Harvey, and Sonnet one on RepLiQA. In the native-reference arm, TechQA Haiku excludes six, and RepLiQA includes 20 answerable items without span gold. MiRAGE Harvey Haiku excludes one failure, leaving 149 items; the other MiRAGE rows contain 150. Excluding failures raises the scores of models that fail, mainly Haiku: generated TechQA Haiku span $F_1$ is 0.40 with exclusion and 0.35 with failures scored zero. These tables are descriptive, not a matched comparison. The repeated comparison in \Cref{sec:focused-results} matches systems across evaluation sets, requires span gold for retrieval scoring, and applies consistent failure handling.

\paragraph{MiRAGE chunk scoring.}
MiRAGE identifies gold by native chunk IDs. Each final predicted span is mapped to the native chunk containing its midpoint, and repeated chunk IDs count once. \Cref{tab:mirage-chunk-retrieval} reports the resulting chunk precision, recall and $F_1$ separately from character-overlap span metrics. With whole-chunk gold, a short quote inside a relevant chunk can have high character precision but recover few gold characters; predictions in non-gold chunks still incur false positives. Chunk scores measure recovery of evidence regions and are not directly comparable to \method's span scores. Different gold granularity and execution budgets preclude attributing cross-table differences to benchmark difficulty alone.

\begin{figure}[H]
\centering
\includegraphics[width=\linewidth]{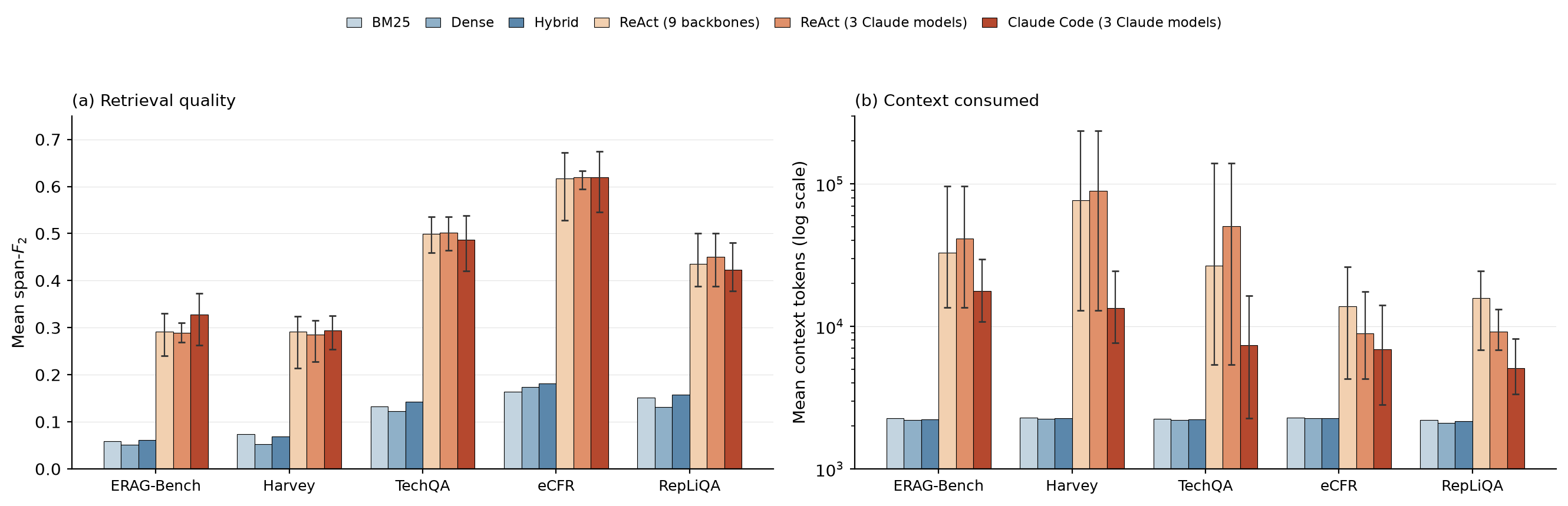}
\caption{Five-corpus system comparison: (a) mean span $F_2$ and (b) mean context tokens, on a logarithmic scale. Agent bars summarize the broader ReAct backbone panel, the three-Claude ReAct panel and the three-Claude Claude Code panel; whiskers show the within-group range. The three-Claude agent groups on TechQA and RepLiQA include negative questions, with credit for correct abstention; the tables and repeated study report retrieval scores on answerable questions.}
\label{fig:five-corpus-measurements}
\end{figure}

\Cref{fig:five-corpus-measurements} separately summarizes the original single-pass comparison of indexed retrievers, ReAct agents and Claude Code agents. Its scoring populations are described in the caption.

\begin{table}[H]
\caption{Retrieval performance on the answerable questions from the five-corpus CoverSynth outputs against a Claude code harness for three differently sized Claude models. File and span retrieval metrics include average precision (P), recall (R), and F1. Resource use is the mean number of tool calls and of evidence-context tokens per question; the turn budget is 50. Failed runs are excluded (Appendix~\ref{app:original-retrieval}).}
\label{tab:downstream}
\begin{center}
{\small
\setlength{\tabcolsep}{4pt}
\begin{tabular}{l l ccc ccc cc}
\toprule
 & & \multicolumn{3}{c}{\textbf{File}} & \multicolumn{3}{c}{\textbf{Span}} & \multicolumn{2}{c}{\textbf{Resource use}} \\
\cmidrule(lr){3-5} \cmidrule(lr){6-8} \cmidrule(lr){9-10}
\textbf{Corpus} & \textbf{Model} & \textbf{P} & \textbf{R} & \textbf{F1} & \textbf{P} & \textbf{R} & \textbf{F1} & \textbf{Tool Calls} & \textbf{Tokens (k)} \\
\midrule
\multirow{3}{*}{ERAG-Bench} & Opus-5    & 0.52 & 0.62 & 0.53 & 0.17 & 0.47 & 0.23 & 13.7 & 11.0 \\
                            & Sonnet-5  & 0.59 & 0.55 & 0.54 & 0.26 & 0.41 & 0.29 & 18.4 & 13.9 \\
                            & Haiku-4.5 & 0.43 & 0.40 & 0.39 & 0.20 & 0.29 & 0.22 & 30.0 & 30.1 \\
\midrule
\multirow{3}{*}{Harvey FK} & Opus-5    & 0.46 & 0.66 & 0.51 & 0.19 & 0.38 & 0.23 & 7.6 & 7.9 \\
                        & Sonnet-5  & 0.50 & 0.54 & 0.49 & 0.20 & 0.29 & 0.22 & 6.7 & 8.7 \\
                        & Haiku-4.5 & 0.42 & 0.40 & 0.38 & 0.23 & 0.19 & 0.19 & 12.3 & 24.0 \\
\midrule
\multirow{3}{*}{TechQA} & Opus-5    & 0.88 & 0.86 & 0.86 & 0.34 & 0.67 & 0.41 & 7.0 & 2.3 \\
                        & Sonnet-5  & 0.82 & 0.77 & 0.79 & 0.35 & 0.53 & 0.39 & 10.6 & 3.5 \\
                        & Haiku-4.5 & 0.79 & 0.73 & 0.75 & 0.43 & 0.45 & 0.40 & 23.8 & 16.1 \\
\midrule
\multirow{3}{*}{eCFR} & Opus-5    & 0.99 & 0.98 & 0.98 & 0.58 & 0.79 & 0.64 & 6.1 & 2.7 \\
                      & Sonnet-5  & 0.96 & 0.93 & 0.94 & 0.71 & 0.63 & 0.63 & 6.7 & 4.1 \\
                      & Haiku-4.5 & 0.91 & 0.89 & 0.90 & 0.72 & 0.57 & 0.60 & 16.5 & 13.2 \\
\midrule
\multirow{3}{*}{RepLiQA} & Opus-5    & 0.67 & 0.74 & 0.68 & 0.38 & 0.51 & 0.41 & 8.4 & 3.7 \\
                         & Sonnet-5  & 0.63 & 0.62 & 0.61 & 0.40 & 0.41 & 0.38 & 12.2 & 3.4 \\
                         & Haiku-4.5 & 0.64 & 0.65 & 0.63 & 0.43 & 0.40 & 0.38 & 18.8 & 7.5 \\
\bottomrule
\end{tabular}
}
\end{center}
\end{table}

\begin{table}[H]
\caption{Retrieval on native ERAG-Bench, TechQA and RepLiQA reference sets using Claude Code. File and span metrics report mean precision (P), recall (R), and $F_1$. Resource use is the mean number of tool calls and evidence-context tokens per question; the turn budget is 40. Failed runs are excluded (\Cref{app:original-retrieval}). ERAG reference gold is document-level, so span metrics are omitted. MiRAGE chunk metrics appear separately in \Cref{tab:mirage-chunk-retrieval}.}
\label{tab:baseline}
\begin{center}
{\small
\setlength{\tabcolsep}{4pt}
\begin{tabular}{l l ccc ccc cc}
\toprule
 & & \multicolumn{3}{c}{\textbf{File}} & \multicolumn{3}{c}{\textbf{Span}} & \multicolumn{2}{c}{\textbf{Resource use}} \\
\cmidrule(lr){3-5} \cmidrule(lr){6-8} \cmidrule(lr){9-10}
\textbf{Corpus} & \textbf{Model} & \textbf{P} & \textbf{R} & \textbf{F1} & \textbf{P} & \textbf{R} & \textbf{F1} & \textbf{Tool Calls} & \textbf{Tokens (k)} \\
\midrule
\multirow{3}{*}{ERAG-Bench} & Opus-5    & 0.95 & 0.88 & 0.90 & --- & --- & --- & 10.8 & 8.8 \\
                            & Sonnet-5  & 0.91 & 0.67 & 0.73 & --- & --- & --- & 9.9 & 7.3 \\
                            & Haiku-4.5 & 0.94 & 0.56 & 0.66 & --- & --- & --- & 10.5 & 10.9 \\
\midrule
\multirow{3}{*}{TechQA} & Opus-5    & 0.83 & 0.91 & 0.86 & 0.23 & 0.72 & 0.32 & 8.1 & 2.5 \\
                        & Sonnet-5  & 0.77 & 0.80 & 0.78 & 0.24 & 0.61 & 0.31 & 11.4 & 4.5 \\
                        & Haiku-4.5 & 0.79 & 0.81 & 0.80 & 0.28 & 0.51 & 0.32 & 17.2 & 13.4 \\
\midrule
\multirow{3}{*}{RepLiQA} & Opus-5    & 0.98 & 0.99 & 0.98 & 0.61 & 0.88 & 0.67 & 3.9 & 1.2 \\
                         & Sonnet-5  & 0.98 & 0.99 & 0.98 & 0.74 & 0.85 & 0.74 & 3.8 & 0.81 \\
                         & Haiku-4.5 & 0.97 & 0.98 & 0.97 & 0.77 & 0.73 & 0.70 & 4.0 & 1.5 \\
\bottomrule
\end{tabular}
}
\end{center}
\end{table}

\begin{table}[H]
\centering
\small
\setlength{\tabcolsep}{3pt}
\caption{MiRAGE retrieval using native chunk gold on Harvey and eCFR. File and chunk precision (P), recall (R) and $F_1$ use the evidence units named in their column groups. Chunk scores are not directly comparable to character-overlap span scores. Claude Code uses a 40-turn budget. $N$ counts successful answerable runs; Harvey Haiku excludes one failure. Resource columns report mean tool calls and evidence-context tokens.}
\label{tab:mirage-chunk-retrieval}
\begin{tabular}{llrccccccrr}
\toprule
 & & & \multicolumn{3}{c}{File} & \multicolumn{3}{c}{Chunk} & \multicolumn{2}{c}{Resource use} \\
\cmidrule(lr){4-6}\cmidrule(lr){7-9}\cmidrule(lr){10-11}
Corpus & Model & $N$ & P & R & $F_1$ & P & R & $F_1$ & Calls & Tokens (k) \\
\midrule
Harvey FK & Opus-5 & 150 & 0.673 & 0.616 & 0.596 & 0.317 & 0.491 & 0.335 & 7.4 & 7.740 \\
 & Sonnet-5 & 150 & 0.621 & 0.485 & 0.498 & 0.350 & 0.365 & 0.310 & 6.3 & 8.489 \\
 & Haiku-4.5 & 149 & 0.636 & 0.455 & 0.486 & 0.411 & 0.316 & 0.309 & 6.4 & 17.632 \\
\midrule
eCFR & Opus-5 & 150 & 0.997 & 0.977 & 0.982 & 0.625 & 0.828 & 0.651 & 5.4 & 2.435 \\
 & Sonnet-5 & 150 & 0.907 & 0.880 & 0.889 & 0.712 & 0.721 & 0.665 & 4.9 & 3.306 \\
 & Haiku-4.5 & 150 & 0.920 & 0.890 & 0.900 & 0.753 & 0.700 & 0.671 & 8.4 & 12.202 \\
\bottomrule
\end{tabular}
\end{table}

\subsection{Focused repeated-study protocol}
\label[appendix]{app:focused-protocol}

\paragraph{Populations and scoring.}
The two evaluation sets for each corpus share the prepared document collection. Generated TechQA and RepLiQA evaluation sets each contribute 100 answerable items with span gold. Human TechQA contributes 150; human RepLiQA contributes 130 after excluding 20 answerable items with no span gold. Negatives are scored separately. Span precision and recall measure character overlap between located predicted and gold spans, pooled across files within an item; span $F_2$ weights recall more than precision. Scores are averaged over repeats, then over items. File recall measures the share of gold documents from which a located snippet appears in the final citations. Failed runs remain in the denominator with score zero. Complete document recall in \Cref{tab:focused-results} requires every system to cite all gold documents in every repeat, not merely to have read a gold document during its trajectory.

\paragraph{Panels.}
The nine-system panel combines BM25, dense and hybrid retrieval with six base agents: ReAct and Claude Code with Haiku-4.5, Sonnet-5 and Opus-5, each run three times with 50 turns and 300\,s per question (\Cref{app:retrieval-config}). The deterministic indexed runs are reused across repeats. Three additional single-run agent configurations are reported separately in \Cref{app:extended-configs} and are excluded from the main analysis.

\begin{table}[H]
\centering
\small
\caption{Focused comparison on answerable items with span gold. The first block uses all nine systems; complete document recall requires every system to cite all gold documents in every repeat. The second block uses the six base agents. A pair meets the detection criterion when nominal paired tests detect a difference in at least 80\% of 400 resamples of 100 items with replacement.}
\label{tab:focused-results}
\setlength{\tabcolsep}{5pt}
\begin{tabular}{lrrrr}
\toprule
 & \multicolumn{2}{c}{TechQA} & \multicolumn{2}{c}{RepLiQA} \\
\cmidrule(lr){2-3}\cmidrule(lr){4-5}
Metric & \method & Human & \method & Human \\
\midrule
\multicolumn{5}{l}{\textit{All nine systems, three repeats}} \\
Complete document recall by all systems & 23/100 & 75/150 & 12/100 & 118/130 \\
Best-system mean span $F_2$ & 0.52 & 0.47 & 0.47 & 0.75 \\
\midrule
\multicolumn{5}{l}{\textit{Six base agents, three repeats}} \\
Span-$F_2$ reliability $G(100)$ & 0.94 & 0.74 & 0.93 & 0.86 \\
Pairs meeting detection criterion & 8/15 & 3/15 & 8/15 & 4/15 \\
\bottomrule
\end{tabular}
\end{table}

\paragraph{Headroom and score interpretation.}
Universal complete document citation occurs on 50\% of human TechQA items versus 23\% of generated items, and on 91\% versus 12\% for RepLiQA (\Cref{tab:focused-results}). The best observed mean span $F_2$ is 0.47 versus 0.75 on RepLiQA but 0.52 versus 0.47 on TechQA (\method\ versus human). Added document-citation headroom therefore does not imply uniformly greater difficulty under every metric. Multiple gold documents and lower lexical overlap are plausible contributors, not isolated causal effects. Document recall concerns final citations, not every document an agent may have read.

The per-system score intervals in \Cref{fig:focused-system-scores} use 1{,}000 bootstrap resamples of the eligible items after averaging each system's repeats. They describe item-sampling uncertainty conditional on those runs; they do not estimate variability across new runs, regenerated questions or new corpora. The two evaluation sets contain different questions, so the displayed means are not a matched-question comparison. These intervals are distinct from the reliability and paired-contrast intervals below.

\paragraph{Reliability and uncertainty.}
Relative generalizability is estimated from a systems $\times$ items $\times$ repeats variance decomposition. Projected to $N=100$ items and $R$ repeats,
\begin{equation}
G(N,R)=\frac{\sigma_s^2}{\sigma_s^2+\sigma_{si}^2/N+\sigma_{sr}^2/R+\sigma_e^2/(NR)}.
\label{eq:ranking-reliability}
\end{equation}
Here $s$, $i$ and $r$ index systems, items and repeats; estimated variance components are truncated at zero. The six-agent analysis uses $R=3$. Confidence intervals resample items and, when present, repeats with replacement. Pooled $\Delta G$ is the arithmetic mean of the TechQA and RepLiQA differences; it does not represent sampling across a larger population of corpora. $G$ is a conditional variance-based reliability coefficient, not a probability that a ranking is correct. Adding the three indexed retrievers to the six base agents drives $G(100)$ to 0.976--0.997 on all four sets, because the retriever--agent gap dominates system variance; at this ceiling $\Delta G$ is slightly negative (\Cref{tab:focused-reliability}). Separating retrievers from agents is therefore not a distinctive property of either set, and the reliability comparison is informative only within the agent panel.

\begin{table}[H]
\centering
\small
\caption{Span-$F_2$ reliability and sensitivity to panel choice. CS = \method; Hu = human. All estimates project to 100 items. Intervals are 95\% bootstrap intervals for $\Delta G=G_{\mathrm{CS}}-G_{\mathrm{Hu}}$; pooled rows average the two corpus differences. The last panel adds BM25, dense, and hybrid retrieval to the six base agents (three repeats).}
\label{tab:focused-reliability}
\begin{tabular}{llcccc}
\toprule
Panel & Corpus & $G_{\mathrm{CS}}$ & $G_{\mathrm{Hu}}$ & $\Delta G$ & 95\% interval \\
\midrule
Six agents, three repeats & TechQA & 0.942 & 0.739 & +0.203 & $[+0.050,+0.659]$ \\
 & RepLiQA & 0.927 & 0.857 & +0.070 & $[-0.056,+0.337]$ \\
 & Pooled & --- & --- & +0.136 & $[+0.041,+0.414]$ \\
\midrule
Six agents + retrievers & TechQA & 0.986 & 0.988 & $-0.002$ & $[-0.011,+0.005]$ \\
 & RepLiQA & 0.976 & 0.997 & $-0.022$ & $[-0.039,-0.016]$ \\
 & Pooled & --- & --- & $-0.012$ & $[-0.022,-0.008]$ \\
\bottomrule
\end{tabular}
\end{table}

\paragraph{Contrasts and pair detection.}
Capability contrasts compare all three pairs of Haiku, Sonnet, and Opus within each of two harnesses on each corpus. Their nominal 95\% intervals bootstrap items and repeats; the 12 intervals per evaluation-set type are correlated and not multiplicity-adjusted. All generated-set contrasts are positive in each repeat. Pair detection is a separate calculation: each system's item score is averaged over repeats, and 400 samples of 100 items are drawn with replacement from the full eligible set. A two-sided paired t-test uses nominal $\alpha=0.05$; a pair meets the criterion if at least 80\% of draws detect a difference. These are estimated detection frequencies. Several pairs lie close to the cutoff: generated RepLiQA ReAct Sonnet--Opus (0.805) and Claude Code Sonnet--Opus (0.84) meet it, whereas generated TechQA ReAct Haiku--Sonnet (0.775), generated TechQA ReAct-Haiku versus Claude Code-Haiku (0.76) and generated RepLiQA Claude Code Haiku--Sonnet (0.755) do not. Pair counts are therefore sensitive to the threshold. The procedure is distinct from the separately computed Holm-adjusted all-pairs significance tests, and nondetection does not establish equivalence.

All 12 within-harness model contrasts have positive nominal 95\% intervals on the generated sets, compared with four positive intervals, one negative interval, and seven crossing zero on the human sets. More six-agent pairs meet the separate detection criterion on the generated sets (\Cref{tab:focused-results}), but no same-model ReAct--Claude Code pair meets it on either generated set (the closest is generated TechQA Haiku at 0.76). These results locate separation in the tested model contrasts; expected model order does not establish benchmark validity, and nondetection of harness differences does not establish harness equivalence.

\paragraph{Detailed scores.}
Tables~\ref{tab:scores-techqa} and~\ref{tab:scores-repliqa} report each system's precision, recall, $F_2$, and cited-document recall. Retriever means average the three indexed methods; base-agent means average the six ReAct/Claude Code configurations. In all four corpus/set combinations, retrievers have higher mean span recall but lower $F_2$ than the base agents: recall is 0.69 versus 0.54 on generated TechQA and 0.96 versus 0.80 on human RepLiQA, for example. Broad returned chunks incur low precision; agents can delimit shorter spans. Citation style also interacts with gold length. A controlled output-budget comparison would be needed to isolate the causal effect of citation length, so $F_2$ should be read alongside precision, recall, and cited-document recall.

\begin{table}[H]
\centering
\footnotesize
\caption{TechQA: mean scores on answerable items with span gold (CS = CoverSynth, Hu = human; n = 100 CS, 150 human). Reps = independent runs over the full evaluation set (not agent turns); agent scores average their repeats, and failed runs score 0. CC-Opus-5.5 and CC-Max are single-run extensions outside the nine-system panel (\Cref{app:extended-configs}).}
\label{tab:scores-techqa}
\begin{tabular}{l c cc cc cc cc}
\toprule
System & Reps & \multicolumn{2}{c}{Span F2} & \multicolumn{2}{c}{Span P} & \multicolumn{2}{c}{Span R} & \multicolumn{2}{c}{File R} \\
\midrule
 &  & CS & Hu & CS & Hu & CS & Hu & CS & Hu \\
\midrule
BM25 & 1 & 0.16 & 0.13 & 0.04 & 0.03 & 0.72 & 0.83 & 0.83 & 0.95 \\
Dense & 1 & 0.15 & 0.12 & 0.04 & 0.03 & 0.61 & 0.77 & 0.72 & 0.95 \\
Hybrid & 1 & 0.17 & 0.12 & 0.05 & 0.03 & 0.75 & 0.82 & 0.84 & 0.97 \\
\midrule
ReAct-Haiku & 3 & 0.39 & 0.47 & 0.39 & 0.31 & 0.43 & 0.64 & 0.65 & 0.86 \\
ReAct-Sonnet & 3 & 0.46 & 0.41 & 0.36 & 0.23 & 0.57 & 0.63 & 0.78 & 0.78 \\
ReAct-Opus & 3 & 0.52 & 0.45 & 0.32 & 0.24 & 0.69 & 0.73 & 0.89 & 0.90 \\
\midrule
CC-Haiku & 3 & 0.34 & 0.41 & 0.37 & 0.30 & 0.37 & 0.53 & 0.63 & 0.83 \\
CC-Sonnet & 3 & 0.46 & 0.40 & 0.38 & 0.24 & 0.53 & 0.59 & 0.76 & 0.78 \\
CC-Opus & 3 & 0.50 & 0.43 & 0.32 & 0.23 & 0.67 & 0.70 & 0.88 & 0.88 \\
CC-Opus-5.5 & 1 & 0.48 & 0.47 & 0.37 & 0.29 & 0.57 & 0.67 & 0.86 & 0.89 \\
\midrule
CC-Max-Haiku & 1 & 0.42 & 0.45 & 0.42 & 0.31 & 0.46 & 0.60 & 0.79 & 0.90 \\
CC-Max-Opus-5.5 & 1 & 0.50 & 0.51 & 0.40 & 0.32 & 0.58 & 0.70 & 0.89 & 0.93 \\
\bottomrule
\end{tabular}
\end{table}

\begin{table}[H]
\centering
\footnotesize
\caption{RepLiQA: mean scores on answerable items with span gold (CS = CoverSynth, Hu = human; n = 100 CS, 130 human). Reps = independent runs over the full evaluation set (not agent turns); agent scores average their repeats, and failed runs score 0. CC-Opus-5.5 and CC-Max are single-run extensions outside the nine-system panel (\Cref{app:extended-configs}).}
\label{tab:scores-repliqa}
\begin{tabular}{l c cc cc cc cc}
\toprule
System & Reps & \multicolumn{2}{c}{Span F2} & \multicolumn{2}{c}{Span P} & \multicolumn{2}{c}{Span R} & \multicolumn{2}{c}{File R} \\
\midrule
 &  & CS & Hu & CS & Hu & CS & Hu & CS & Hu \\
\midrule
BM25 & 1 & 0.19 & 0.15 & 0.06 & 0.03 & 0.56 & 0.98 & 0.71 & 0.99 \\
Dense & 1 & 0.16 & 0.15 & 0.05 & 0.03 & 0.49 & 0.91 & 0.69 & 0.99 \\
Hybrid & 1 & 0.20 & 0.16 & 0.06 & 0.04 & 0.58 & 0.99 & 0.78 & 1.00 \\
\midrule
ReAct-Haiku & 3 & 0.36 & 0.70 & 0.35 & 0.72 & 0.39 & 0.75 & 0.59 & 0.98 \\
ReAct-Sonnet & 3 & 0.42 & 0.74 & 0.40 & 0.70 & 0.45 & 0.82 & 0.69 & 0.99 \\
ReAct-Opus & 3 & 0.47 & 0.73 & 0.35 & 0.54 & 0.57 & 0.87 & 0.82 & 0.98 \\
\midrule
CC-Haiku & 3 & 0.32 & 0.66 & 0.37 & 0.73 & 0.33 & 0.69 & 0.57 & 0.98 \\
CC-Sonnet & 3 & 0.38 & 0.75 & 0.42 & 0.72 & 0.39 & 0.81 & 0.63 & 0.99 \\
CC-Opus & 3 & 0.44 & 0.73 & 0.36 & 0.56 & 0.49 & 0.85 & 0.77 & 0.99 \\
CC-Opus-5.5 & 1 & 0.42 & 0.73 & 0.45 & 0.74 & 0.44 & 0.77 & 0.75 & 1.00 \\
\midrule
CC-Max-Haiku & 1 & 0.35 & 0.62 & 0.37 & 0.67 & 0.37 & 0.67 & 0.70 & 0.99 \\
CC-Max-Opus-5.5 & 1 & 0.46 & 0.70 & 0.47 & 0.67 & 0.48 & 0.75 & 0.79 & 1.00 \\
\bottomrule
\end{tabular}
\end{table}

\subsection{Repeat variability and failure robustness}
\label{app:focused-robustness}

The standard deviation of each base system's mean across repeats ranges from 0.002 to 0.022. Removing every item on which any base agent failed retains 66 generated and 128 human TechQA items, and 74 generated and 130 human RepLiQA items. The resulting $G(100)$ point estimates remain higher on the generated sets: 0.91 versus 0.74 on TechQA and 0.92 versus 0.86 on RepLiQA. These are projections to 100 items from the retained populations, not additional significance results. Repeating retrieval evaluation does not measure variability from regenerating the questions or choosing new corpora.

\paragraph{Corpus preparation and prospective uses.}
Corpus preparation uses existing gold documents where available, although generation does not require expert-authored questions as input. The experiments do not isolate the contribution of closed-book answerability. Continuous regeneration and training-data applications remain prospective uses; neither a corpus-update benefit nor a training gain is measured here.

\subsection{Single-run agent extensions}
\label[appendix]{app:extended-configs}

Three further agent configurations were run once on the two-corpus study. They are excluded from the main panel because they have one repeat instead of three and, for CC-Max, a different prompt, architecture, and budget. \emph{CC-Opus-5.5} is the base Claude Code agent with Opus-5.5. \emph{CC-Max} is a Claude Code orchestrator that can only delegate: a hook refuses its own file-reading and search tools, so retrieval is performed by scout, reader, and verifier subagents of the same model (Haiku-4.5 or Opus-5.5), guided by a retrieval playbook, with 100 turns and 900\,s per question. Answer and abstention instructions match the base agents. Scores appear in Tables~\ref{tab:scores-techqa} and~\ref{tab:scores-repliqa}. CC-Opus-5.5 does not exceed CC-Opus-5 on either generated set (0.48 versus 0.50 span $F_2$ on TechQA, 0.42 versus 0.44 on RepLiQA).

Adding the three extensions to the six base agents requires averaging repeats per system and item and using a systems $\times$ items decomposition. The reliability advantage then shrinks: $G(100)$ is 0.916 versus 0.792 on TechQA ($\Delta G=+0.124$, 95\% interval $[+0.009,+0.239]$) and 0.911 versus 0.903 on RepLiQA ($+0.008$, $[-0.063,+0.085]$), and the pooled difference, $+0.066$ $[-0.001,+0.135]$, includes zero.

CC-Max illustrates a reducible failure mode under a changed prompt, retrieval procedure and budget. On answerable generated items, base CC-Haiku ends without an answerable verdict and without citing a gold document in 82/300 TechQA runs and 66/300 RepLiQA runs, compared with 9/100 and 8/100 for CC-Max-Haiku. Base turn-cap failures account for 38/300 and 32/300 runs; CC-Max has three timeout failures per 100 runs on each corpus. Because the base has three repeats and CC-Max one, with differing instructions and budgets, this does not isolate an architecture effect.

\subsection{Negative-query results and original-figure conventions}
\label[appendix]{app:negative-results}

\Cref{tab:negatives} reports the answerability-verdict accuracy for a focused analysis of TechQA and RepLiQA, counting failures as incorrect. The generated families contain only 2--13 items per corpus (\Cref{tab:realized-structure}), so rates can move substantially when a small number of judgments change. These results concern verdict agreement, not the quality of a follow-up clarification or premise correction.

The heatmap (\Cref{fig:negatives}) uses single-pass Claude Code summaries for all five corpora and excludes failed runs in every cell. Its displayed values must therefore not be combined with the repeated, failure-inclusive table as though they shared one denominator. Correct abstention on an empty-gold negative can receive a span score of one for an agent, whereas the retrievers do not return an answerability verdict. This is an additional reason that the agent--retriever $F_2$ gap cannot be interpreted solely as better evidence retrieval.

\begin{table}[H]
\centering
\scriptsize
\caption{Answerability-verdict accuracy on negatives, by failure family (CoverSynth) and on the human 'not in corpus' negatives. Failed runs count as wrong; agents averaged over repeats. CC-Opus-5.5 and CC-Max are single-run extensions (\Cref{app:extended-configs}).}
\label{tab:negatives}
\resizebox{\linewidth}{!}{%
\begin{tabular}{l ccccc ccccc}
\toprule
 & \multicolumn{5}{c}{TechQA} & \multicolumn{5}{c}{RepLiQA} \\
\midrule
System & Unsupp. & False prem. & Underspec. & Ambig. & Hu unsupp. & Unsupp. & False prem. & Underspec. & Ambig. & Hu unsupp. \\
n items & 12 & 9 & 5 & 2 & 120 & 13 & 9 & 12 & 13 & 38 \\
\midrule
ReAct-Haiku & 0.67 & 1.00 & 0.00 & 0.00 & 0.54 & 0.79 & 0.96 & 0.06 & 0.08 & 0.35 \\
ReAct-Sonnet & 0.75 & 0.89 & 0.00 & 0.00 & 0.79 & 0.95 & 0.85 & 0.14 & 0.21 & 0.39 \\
ReAct-Opus & 0.81 & 1.00 & 0.00 & 0.00 & 0.71 & 0.97 & 1.00 & 0.25 & 0.64 & 0.43 \\
CC-Haiku & 0.81 & 0.85 & 0.13 & 0.00 & 0.66 & 0.72 & 0.85 & 0.06 & 0.05 & 0.26 \\
CC-Sonnet & 0.83 & 0.96 & 0.00 & 0.00 & 0.83 & 0.92 & 1.00 & 0.31 & 0.36 & 0.45 \\
CC-Opus & 0.89 & 1.00 & 0.13 & 0.00 & 0.83 & 0.95 & 1.00 & 0.42 & 0.62 & 0.47 \\
CC-Opus-5.5 & 0.92 & 1.00 & 0.00 & 0.50 & 0.78 & 0.85 & 1.00 & 0.33 & 0.69 & 0.34 \\
CC-Max-Haiku & 0.75 & 0.78 & 0.00 & 0.00 & 0.60 & 0.69 & 1.00 & 0.08 & 0.23 & 0.21 \\
CC-Max-Opus-5.5 & 0.92 & 1.00 & 0.00 & 0.00 & 0.78 & 0.92 & 1.00 & 0.17 & 0.46 & 0.45 \\
\bottomrule
\end{tabular}
}
\end{table}


\end{document}